\documentclass[10pt,twocolumn,letterpaper]{article}

\usepackage{iccv}
\usepackage{times}
\usepackage{epsfig}
\usepackage{graphicx}
\usepackage{amsmath}
\usepackage{amssymb}

\usepackage{xcolor}
\usepackage{xspace}
\usepackage{algorithm}
\usepackage{algorithmic}
\usepackage{booktabs} 
\usepackage{multirow}
\usepackage{amsthm}
\usepackage{pifont}%
\newcommand{\cmark}{\ding{52}}%
\newcommand{\xmark}{\ding{56}}%
\usepackage{listings}

\newtheorem{theorem}{Theorem}
\newtheorem{lemma}{Lemma}

\usepackage[pagebackref=true,breaklinks=true,letterpaper=true,colorlinks,bookmarks=false]{hyperref}

\DeclareMathOperator{\Tr}{Tr}
\newcommand{\ourmodel}{Differentiable Mask-Matching Network} 
\newcommand{\ourmodelshort}{DMM-Net}

\makeatletter
\DeclareRobustCommand\onedot{\futurelet\@let@token\@onedot}
\def\@onedot{\ifx\@let@token.\else.\null\fi\xspace}

\def\eg{\emph{e.g}\onedot} 
\def\ie{\emph{i.e}\onedot}

\def\wrt{w.r.t\onedot} 

\makeatother

\definecolor{codegreen}{rgb}{0,0.6,0}
\definecolor{codegray}{rgb}{0.5,0.5,0.5}
\definecolor{codepurple}{rgb}{0.58,0,0.82}
\definecolor{backcolour}{rgb}{0.95,0.95,0.92}

\lstdefinestyle{mystyle}{
    backgroundcolor=\color{backcolour},   
    commentstyle=\color{codegreen},
    keywordstyle=\color{magenta},
    numberstyle=\tiny\color{codegray},
    stringstyle=\color{codepurple},
    basicstyle=\footnotesize,
    breakatwhitespace=false,         
    breaklines=true,                 
    captionpos=b,                    
    keepspaces=true,                 
    numbers=left,                    
    numbersep=5pt,                  
    showspaces=false,                
    showstringspaces=false,
    showtabs=false,                  
    tabsize=2
}
 
\iccvfinalcopy 

\def\iccvPaperID{2725} 

\ificcvfinal\fi
\begin{document}

\title{DMM-Net: Differentiable Mask-Matching Network \\ for Video Object Segmentation}

\author{Xiaohui Zeng\textsuperscript{1, 2}\thanks{Equal contribution.} \qquad Renjie Liao\textsuperscript{1, 2, 3*} \qquad Li Gu\textsuperscript{1} \qquad Yuwen Xiong\textsuperscript{1, 2, 3} \\ \qquad Sanja Fidler\textsuperscript{1, 2, 4} \qquad Raquel Urtasun\textsuperscript{1, 2, 3, 5}\\
University of Toronto\textsuperscript{1} \quad Vector Institute\textsuperscript{2} \quad Uber ATG Toronto\textsuperscript{3} \\ 
NVIDIA\textsuperscript{4} \quad Canadian Institute for Advanced Research\textsuperscript{5}\\
{\tt\small \{xiaohui, rjliao, yuwen, fidler\}@cs.toronto.edu} \quad {\tt\small li.gu@mail.utoronto.ca} \quad {\tt\small urtasun@uber.com}
}

\maketitle

\begin{abstract}
In this paper, we propose the differentiable mask-matching network (DMM-Net) for solving the video object segmentation problem where the initial object masks are provided.
Relying on the Mask R-CNN backbone, we extract mask proposals per frame and formulate the matching between object templates and proposals at one time step as a linear assignment problem where the 
cost matrix is predicted by a CNN.
We propose a differentiable matching layer by unrolling a projected gradient descent algorithm in which the projection exploits the Dykstra's algorithm.
We prove that under mild conditions, the matching is guaranteed to converge to the optimum.
In practice, it performs similarly to the Hungarian algorithm during inference. 
Meanwhile, we can back-propagate through it to learn the cost matrix.
After matching, a refinement head is leveraged to improve the quality of the matched mask. 
Our DMM-Net achieves competitive results on the largest video object segmentation dataset YouTube-VOS.  
On DAVIS 2017, DMM-Net achieves the best performance without online learning on the first frames.
Without any fine-tuning, DMM-Net performs comparably to state-of-the-art methods on SegTrack v2 dataset.
At last, our matching layer is very simple to implement;
we attach the PyTorch code ($<50$ lines) in the supplementary material. Our code is released at \url{https://github.com/ZENGXH/DMM_Net}.
\end{abstract}


\vspace{-2mm}
\section{Introduction}

Video object and instance segmentation problems have received significant attention~\cite{badrinarayanan2010label,tsai2012motion, maninis2018video} attributed to the recent availability of high-quality datasets, \eg, YouTube-VOS~\cite{xu2018youtube}, DAVIS~\cite{perazzi2016benchmark,pont2017}.
Given an input video, the aim is to separate the objects or instances from the background at the pixel-level.
This is a fundamental computer vision task due to its wide range of applications including autonomous driving, video surveillance, and video editing.

\begin{figure}[t]
	\centering
	\begin{tabular}{@{\hspace{0mm}}c@{\hspace{0 mm}}c@{\hspace{0 mm}}c@{\hspace{0mm}}c@{\hspace{0mm}}c}
		\includegraphics[width=0.495\linewidth]{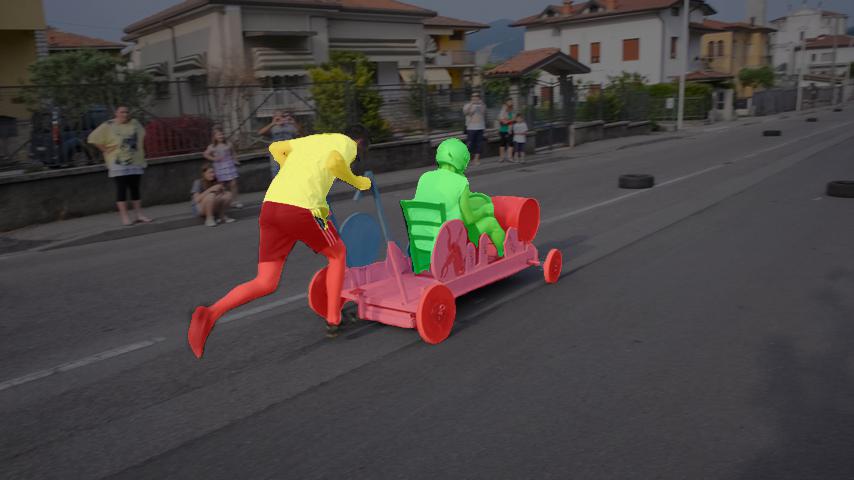}&
		\includegraphics[width=0.495\linewidth]{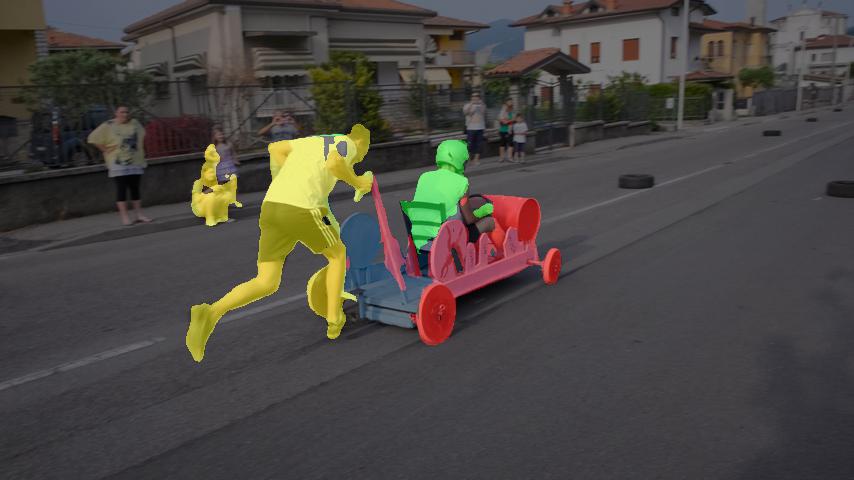}\\
		{\footnotesize (a) RGMP~\cite{wug2018fast} } &{\footnotesize (b) CINM~\cite{bao2018cnn}} \\
		\includegraphics[width=0.495\linewidth]{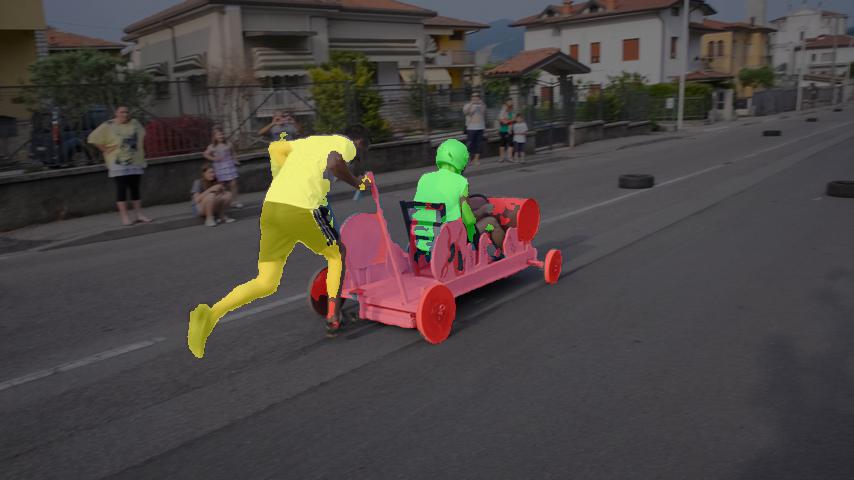}&
		\includegraphics[width=0.495\linewidth]{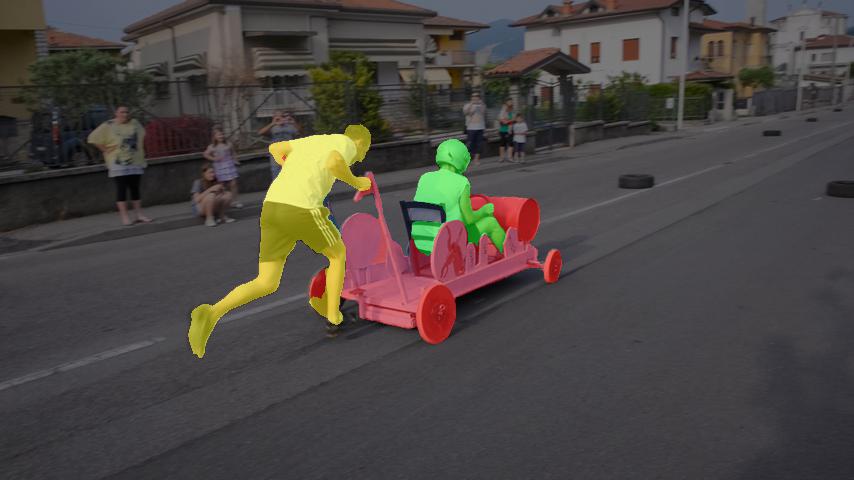}\\
		{\footnotesize (c) PReMVOS~\cite{luiten2018premvos} } &{\footnotesize (d) Ours} \\[0.7mm]
	\end{tabular}
	\caption{Visual comparison with some competitive methods on the last frame of the \textit{soapbox} sequence of DAVIS 2017 dataset. Even though video segmentation quality generally degrades as time goes on, our model still provides better details.} 
	\vspace{-0.5cm}
	\label{fig:vis_comp}
\end{figure}

Two main setups of this problem are \textit{unsupervised} and \textit{semi-supervised} which differ from each other in whether the ground-truth annotated masks of the object  instances in the first frame of the video are provided~\cite{perazzi2016benchmark,pont2017} during inference.
In this paper, we focus on the \textit{semi-supervised} setting, \ie, the instance masks are provided for the first frames of the test videos. 
However, even with some annotated information at test time, this task is still very challenging. 
For example, the algorithm needs to deal with not only the dramatic appearance changes, occlusion and deformation but also potentially with large camera and object motion.
Furthermore, the expectation from a good video instance segmentation model is to produce temporally cohesive and stable predictions, which presents an additional challenge.

Existing algorithms typically leverage pretrained deep neural networks to predict object instance masks.
Some of them, \eg,~\cite{maninis2018video}, directly predict the masks in a frame-independent way and achieve surprisingly good performance which verifies the great transfer ability of deep neural networks.
Many algorithms leverage the previously predicted masks in various ways thus enabling the mask propagation over time.
This strategy is demonstrated to improve temporal coherence and segmentation quality.
Moreover, template matching between the reference and current frames is often exploited at pixel or mask level to deal with the object disappearance-reappearance phenomenon, occlusion, and fast motion.
However, to the best of our knowledge, none of the existing work integrates the optimal matching algorithm into their framework, which may partly be due to the non-differentiable nature of the problem.

In this paper, we propose the differentiable mask-matching network ({\ourmodelshort}).
We first extract mask proposals via a pretrained Mask R-CNN~\cite{he2017mask} in a frame-independent manner.
For each time step, we then match proposals with the templates in the reference frame such that at most one mask proposal is assigned to one template instance.
The matching cost between a pair of the template and proposal masks is determined based on the intersection-over-union (IoU) of masks and the cosine similarity between their feature maps predicted by a deep convolutional neural network (CNN). 
The key contribution of our paper is that we introduce a differentiable matching layer which solves the linear assignment problem.

Specifically, we unroll the projected gradient descent algorithm in which the challenging projection step is achieved by an efficient cyclic projection method known as \textit{Dykstra}'s algorithm. 
The proposed double-loop matching algorithm is very simple to implement, guaranteed to converge, and achieves similar performance to the optimal matching obtained by the Hungarian algorithm~\cite{kuhn1955hungarian,munkres1957algorithms}.
More importantly, it is fully differentiable which enables learning of the matching cost, thus having a better chance of handling large deformation and appearance changes. 
After matching, we adopt a refine head to refine the matched mask.
Note that our main contribution is somewhat orthogonal to many existing work in a way that our differentiable matching can be integrated with other networks to potentially boost their performance.
On DAVIS 2017~\cite{pont2017}, SegTrack v2~\cite{li2013video} and YouTube-VOS~\cite{xu2018youtube} datasets, our model achieves either state-of-the-art or comparable performance.

\section{Related Work}
The problem of video object/instance segmentation has been extensively studied in the past~\cite{tsai2012motion,badrinarayanan2010label, perazzi2016benchmark, pont2017}.
Many algorithms in this field rely on techniques like template matching which are popular in the object tracking and image matching literature~\cite{yilmaz2006object,lowe2004distinctive,bertinetto2016fully,revaud2016deepmatching}.
However, video object/instance segmentation is more challenging than tracking as it requires pixel-level object/instance mask as output rather than the bounding box.
Meanwhile, it is also very different from matching in that it requires semantic understanding of the image rather than the similarities of low-level cues like color, motion and so on.
Related literature can be classified based on the problem setup, \ie, \textit{unsupervised} vs. \textit{semi-supervised}.
Methods under the \textit{unsupervised} category~\cite{brox2010object,grundmann2010efficient,wang2015saliency,tokmakov2017learning} typically exploit the dense optical flow and appearance feature to group the pixels within the spatio-temporal neighborhood.
%
%
%
In this paper, we focus on the \textit{semi-supervised} setting. 
Based on whether an explicit matching between template and proposal mask is performed and at which level the matching is performed, we can further divide the related work into three sub-categories.

\paragraph{Pixel-level Matching}
Pixel-level matching network (PLM)~\cite{shin2017pixel} first exploits a Siamese type of CNN to extract features of the current frame and the masked reference frame. 
Based on the features, it computes the pixel-level similarity scores and the instance masks.
VideoMatch~\cite{hu2018videomatch} applies a CNN to extract features from the reference and the current frames, respectively.
The feature of the reference frame is further split into the foreground and background ones which are used to compute the similarity with the feature of the current frame via a specially designed soft-matching module.
The similarity-weighted combination of feature is used to predict the final mask.
A fully convolutional Siamese network based approach (SiamMask) is proposed in~\cite{wang2018fast}.
It computes the depth-wise cross correlation between features of templates in the reference and the current frames.
It also consists of mask, box, and score prediction heads similar to Mask R-CNN.
Although these methods do not solve the exact matching problem, the pixel-level similarity scores output by a learnable CNN are still helpful for the task of mask prediction.
However, since cross-correlation between different pixels from the template and the current frame is required, they tend to be intensive on computation and memory. 

\paragraph{Mask-level Matching}
Instead of operating on the pixel-level, some methods including our {\ourmodelshort} resort to the mask-level matching.
Based on pre-computed feature maps, DyeNet~\cite{li2018video} iteratively uses the re-identification and the recurrent mask propagation modules to retrieve disappearing-reappearing objects and handle temporal variations of pose and scale separately. 
Authors in~\cite{cheng2018fast} propose to track the object parts in the video and also compute the similarity scores between the proposal and template parts in the reference frame in order to deal with the missing of tracking and background noises.
In these work, matching is computationally light due to the number of masks/parts is significantly smaller than the number of pixels.
However, they all exploit the greedy solution for matching, \ie, for each template, it returns the maximum-scored assignment if the score is above some threshold otherwise returns no assignment.
In contrast, we solve the matching problem via an iterative solver which is better than the greedy solution in most cases as verified by the experiments.

\paragraph{No Explicit Matching}
Some of the recent works directly exploit deep neural networks to predict the masks.
In~\cite{caelles2017one,maninis2018video}, a pretrained CNN is first fine-tuned to predict both the segmentation mask and contour per frame and then a boundary snapping step is applied to combine both results. 
Authors in~\cite{voigtlaender2017online} later extended this work by introducing an online adaptation step to bootstrap the foreground object segmentation.
Video propagation network (VPN)~\cite{jampani2017video} proposes a bilateral network along with a CNN to propagate the previously predicted masks and images.
MaskRNN~\cite{hu2017maskrnn} exploits the optical flow, images and mask proposals in a recurrent fashion to predict the masks per frame.
MaskTrack~\cite{perazzi2017learning} uses a CNN which takes the last predicted instance mask and the current frame as input and outputs the refined mask. 
Spatial propagation network (SPN)~\cite{cheng2017learning} performs foreground segmentation and instance recognition simultaneously and then refines the instance masks using a spatial propagation module.
Pixel-wise metric learning (PML)~\cite{chen2018blazingly} formulates the video object segmentation as a pixel-wise retrieval problem where the embedding space is predicted by a CNN and then learned via triplet-constrained metric learning.
Based on optical flow and a spatial CNN, a pixel-level spatio-temporal Markov random field (MRF) is built in~\cite{bao2018cnn} where approximate inference is achieved by a CNN.
Authors in~\cite{Yang2018EfficientVO} propose two sub-networks to compute the visual feature of the templates and spatial attention map from the last frame respectively to guide the mask prediction.
Relying on a U-Net, authors in~\cite{wug2018fast} combine the concatenation of the current frame with last predicted mask and the concatenation of the reference frame with the template mask to predict the current mask.
These works are orthogonal to ours in the sense that we can use some of their networks as our feature extractor, while our matching layer could also potentially improve their models.

\section{Model}

\begin{figure*}[h]
    \centering
    \includegraphics[width=0.95\textwidth]{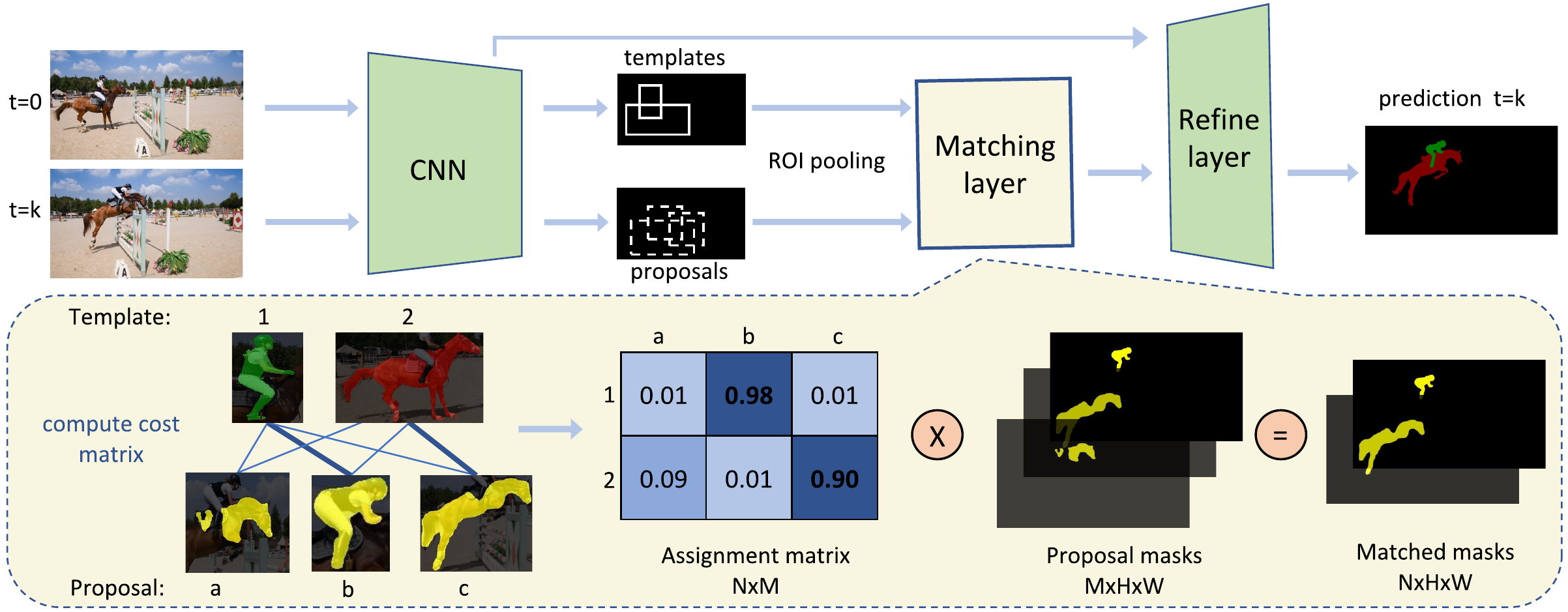}
    \caption{The overall architecture of our model. The yellow box in the bottom indicates the process of bipartite matching and the corresponding output masks as described in the Eq.~(\ref{eq:output_mask}).}
    \label{fig:model}
\end{figure*}

In this section, we introduce our approach which consists of two key components: differentiable mask matching and mask refinement.
Our model assumes we have access to mask proposals in each frame. 
We first explain how we obtain the mask proposals. 
Then we describe our differentiable mask matching approach and discuss the mask refinement. 
Overview of our approach is illustrated in Fig.~\ref{fig:model}.

We assume a video has $T$ frames. 
The mask templates in the first frame are denoted as $R = \{ r_{i} \vert i = 1, \dots, n \}$ where $n$ is the total number of instances throughout the video. 

\paragraph{Mask Proposal Generation}

We first extract mask proposals independently per frame with a COCO-pretrained Mask R-CNN~\cite{he2017mask} (details in Sec.~\ref{sect:implement}).  
We only keep the top-$50$ proposals based on their scores, ensuring that recall is sufficiently high. This step is performed off-line, \ie, our method will run on top of these fixed proposals. 
We denote mask proposals in frame $t$ as $P^{t} = \{ p_{j}^{t} \vert j = 1, \dots, m^{t} \}$ where $m^{t}$ is the total number of proposals at time $t$. 


\paragraph{Differentiable Mask Matching}

\label{sec:matching}

The main motivation for performing object-level matching is to deal with the cases where large deformation, motion and dramatic appearance change are present. 
As aforementioned, proposal based matching is typically superior to optical flow based methods when motion is large. 
Moreover, we design a learnable matching cost which could potentially handle the dramatic appearance change and deformation. 

In particular, at time step $t$, we use a CNN, denoted as $f_{\theta}$, to extract features for the mask proposals $P^{t}$ and the templates $R$ in the first frame. Here $\theta$ denotes the learnable parameters.
The details of the feature extractor is explained in Sec.~\ref{sect:implement}. 
For the $i$-th mask template $r_{i}$ (ground-truth mask in first frame) and the $j$-th mask proposal $p_{j}^{t}$, we compute their features as $f_{\theta}(r_{i})$ and $f_{\theta}(p_{j}^{t})$, respectively.
The matching cost matrix $C^{t}$ consists of the cosine similarity between features and IoU between masks as below,
{
\begin{align}\label{eq:cost_matrix}
	C^{t}_{i, j} = (\lambda - 1) \cos{ \left( f_{\theta}(p_{j}^{t}), f_{\theta}(r_{i}) \right) } - \lambda \text{IoU} (p_{j}^{t}, r_{i}), 
\end{align}
}%
where $\lambda$ is a hyperparameter and $0 < \lambda < 1$.
The overall cost matrix $C^{t}$ is of size $n \times m^{t}$ where each row and column correspond to a template and a mask proposal, respectively. From here on out, we drop the superscript $t$ for simplicity.

\begin{algorithm}[t]
\caption{: Projected Gradient Descent for Matching}\label{alg:matching}
\begin{algorithmic}[1]
\STATE \textbf{Input:} $N_{\text{grad}}, N_{\text{proj}}, X, \alpha, C$ 
\STATE \textbf{Initialization:} $X^{0} = X$
\STATE \textbf{For} $i = 1, 2, \dots, N_{\text{grad}}$:
\STATE \qquad $X^{i} = X^{i-1} - \alpha C$
\STATE \qquad $Y_{0} = X^{i}, q_{1} = 0, q_{2} = 0, q_{3} = 0$
\STATE \qquad \textbf{For} $j = 1, 2, \dots, N_{\text{proj}}$:
\STATE \qquad \qquad $Y_{1} = \mathcal{P}_{1}( Y_{0} + q_{1} )$, \quad $q_{1} = Y_{0} + q_{1} - Y_{1}$
\STATE \qquad \qquad $Y_{2} = \mathcal{P}_{2}( Y_{1} + q_{2} )$, \quad $q_{2} = Y_{1} + q_{2} - Y_{2}$
\STATE \qquad \qquad $Y_{3} = \mathcal{P}_{3}( Y_{2} + q_{3} )$, \quad $q_{3} = Y_{2} + q_{3} - Y_{3}$
\STATE
\STATE \qquad $X^{i} = Y_{3}$
\STATE Return $\hat{X} = \frac{1}{N_{\text{grad}}} \sum_{i=1}^{N_{\text{grad}}} X^{i}$
\end{algorithmic}
\end{algorithm}

We now introduce how we solve the bipartite matching problem.
In particular, we first formulate the minimum-cost bipartite matching as the following integer linear programming (ILP) problem,
{
\begin{align}\label{eq:ILP}
	\min_{X} \qquad & \Tr{\left( CX^{\top} \right)}  \nonumber \\
	s.t. \qquad & X \mathbf{1}_{m} = \mathbf{1}_{n} \nonumber \\
	& X^{\top} \mathbf{1}_{n} \le \mathbf{1}_{m} \nonumber \\
	& X \ge 0 \nonumber \\
	& X_{i, j} \in \{0, 1\} \qquad \forall (i, j)
\end{align}
}%
where $X \in \mathbb{R}^{n \times m}$ is the boolean assignment matrix. 
$\mathbf{1}_{n}$ and $\mathbf{1}_{m}$ are all one vectors with size $n$ and $m$, respectively.
Here we slightly abuse the notation such that subscript $i,j$ denotes the $(i, j)$-th entry of the matrix.
We add the constraint $X \ge 0$ which will be helpful in understanding the relaxed version introduced later.
Note that the problem formulated in Eq.~(\ref{eq:ILP}) and the standard linear assignment problem (LAP) are slightly different in that we replace $X^{\top} \mathbf{1} = \mathbf{1}$ with $X^{\top} \mathbf{1} \le \mathbf{1}$.
This is due to the fact that the number of proposals $m$ is much larger than the number of templates $n$ in our case, \ie, $X$ is a wide matrix.

To solve this ILP problem, one can introduce dummy variables to make $X$ a squared matrix and then use the Hungarian method to optimize  the standard LAP.
However, this naive extension increases the time complexity to $O(m^{3})$ and is not easy to back-propagate through.
Also, we may not necessarily require the exact matching, \ie, real-valued approximated assignment matrix $X$ may be sufficient for the later stage. 
Therefore, we resort to the following linear programming (LP) relaxation,
{
\begin{align}\label{eq:LP}
	\min_{X} \qquad & \Tr{\left( CX^{\top} \right)}  \nonumber \\
	s.t. \qquad & X \mathbf{1}_{m} = \mathbf{1}_{n} \nonumber \\
	& X^{\top} \mathbf{1}_{n} \le \mathbf{1}_{m} \nonumber \\
	& X \ge 0.
\end{align}
}%
Although there are many standard solvers for LP, \eg, the simplex method and the interior point methods, we here introduce a differentiable and easy-to-implement projected gradient descent algorithm.
The algorithm is presented in Alg.~\ref{alg:matching} where $N_{\text{grad}}, N_{\text{proj}}$ are the number of gradient decent (outer-loop) steps and the number of projection (inner-loop) steps, respectively.

At each iteration, we update $X$ following the negative gradient direction.
The major challenge lies in projecting the updated $X$ onto the constraint set.
It is not an easy task since the constraint set in Eq.~(\ref{eq:LP}) is the intersection between three closed convex sets.

To compute the projection, we adopt a cyclic constraint projection method, known as \textit{Dykstra's algorithm}~\cite{dykstra1983algorithm,boyle1986method} which is proved to be convergent for projection onto the non-empty intersection of finite closed convex sets.
The key idea is to break the whole constraint set into multiple simple subsets such that we can easily find the projection operator.
In particular, we can split the constraint set $\mathcal{C}$ into individual constraints, \ie, $\mathcal{C} = \mathcal{C}_{1} \bigcap \mathcal{C}_{2} \bigcap \mathcal{C}_{3}$ where
{
\begin{align}\label{eq:constraint_decomposition}
	\mathcal{C}_{1} & = \{ X \vert X \mathbf{1}_{m} = \mathbf{1}_{n} \} \nonumber \\	
	\mathcal{C}_{2} & = \{ X \vert X^{\top} \mathbf{1}_{n} \le \mathbf{1}_{m} \} \nonumber \\
	\mathcal{C}_{3} & = \{ X \vert X \ge 0 \}. 
\end{align}
}%
It is straightforward to derive the projection operators \wrt each constraint as follows,
{
\begin{align}\label{eq:constraint_projection_1}
	\mathcal{P}_{1}(X) & = X - \frac{1}{m} (X \mathbf{1}_{m} - \mathbf{1}_{n}) \mathbf{1}_{m}^{\top}
\end{align}
\begin{align}\label{eq:constraint_projection_2}
	\mathcal{P}_{2}(X) = \left\{ \begin{array}{ll} 
							X \qquad & \text{if}~~~ X^{\top} \mathbf{1}_{n} \le \mathbf{1}_{m} \\
							X - \frac{1}{n} \mathbf{1}_{n} (\mathbf{1}_{n}^{\top} X - \mathbf{1}_{m}^{\top})  & \text{otherwise}
							\end{array} \right.							
\end{align}
\begin{align}\label{eq:constraint_projection_3}
    \mathcal{P}_{3}(X) & = X^{+} 
\end{align}
}%
Note that $\mathcal{P}_{3}$ is just the ReLU operator.
All these projection operators are differentiable and simple.
\textit{Dykstra's algorithm} works by iteratively projecting the corrected point onto individual constraint set in a cyclic order and then updating the correction by the difference between pre-projection and post-projection. 
The final solution is obtained by averaging the intermediate projected assignment matrices.

The convergence results of the Dykstra's algorithm are established in~\cite{dykstra1983algorithm,boyle1986method,deutsch1994rate}.
Relying on the convergence analysis under the framework of inexact projection primal first-order methods for convex optimization in~\cite{patrascu2018convergence}, we derive the following convergence result of our projected gradient descent algorithm for matching.
\begin{theorem}\label{them:converge}
Let $r_{0} = \Vert X^{0} - X^{\ast} \Vert_{F}$ where $X^{0}$ and $X^{\ast}$ are the initial and optimal assignment matrices, respectively. 
Let the learning rate $0 < \alpha < \min (15 r_{0}, r_{0}/\Vert C \Vert_{F})$.
There exists some constants $0 \le c < 1$ and $\rho > 0$ such that at any iteration of outer loop $i$ in Alg.~\ref{alg:matching}, the error of projection $\Vert X^{i} - \mathcal{P}_{\mathcal{C}}(X^{i}) \Vert_{F} \le \delta = \rho c^{N_{\text{proj}}}$ where $X^{i}$ and $\mathcal{P}_{\mathcal{C}}(X^{i})$ are the assignment matrix and its correct projection onto $\mathcal{C}$, respectively.
Moreover, for any $0 < \epsilon < 1$, there exists a $N_{\text{proj}} \ge \log_{1/c} \left( \rho  \sqrt{\frac{15K}{\alpha \epsilon}} \right)$ such that,
\begin{align}
    \delta \le \frac{\alpha \epsilon }{15 r_{0}} \nonumber
\end{align}
and after at most $K$ iterations where
\begin{align}
    K = \left\lceil {\frac{6r_{0}^{2}}{\alpha \epsilon}} \right\rceil, \nonumber
\end{align}
the output of Alg.~\ref{alg:matching} $\hat{X}$ is $\epsilon$-optimal, \ie, $\Vert \hat{X} - \mathcal{P}_{\mathcal{C}}(\hat{X}) \Vert_{F} \le \epsilon$ and $\vert \Tr{\left( C \hat{X}^{\top} \right)} - \Tr{\left( CX^{\ast\top} \right)} \vert \le \epsilon$.
\end{theorem}
We leave the proof to the supplementary material.
In practice, the convergence is typically observed with moderately large $N_{\text{proj}}$ and reasonable learning rate. 
The implementation of the overall algorithm is very simple.
Please see an example implementation using PyTorch (less than $50$ lines) and empirical convergence analysis with different hyperparameters in the supplementary material.



\paragraph{Mask Refinement}
After matching, for each template, we need to output one mask which is fed to the refinement stage.
Recall that we obtain the optimal assignment $\hat{X}$ (approximately optimal if the previous algorithm does not run till convergence), we can compute a weighted combination of the proposal masks $P$.
Specifically, we first resize mask proposals such that they have the same resolution as the input image.
We then paste them into void images to get the full masks which have the same size as the input images, denoted as $\tilde{P}$.
We obtain the matched mask $\hat{P}$ as:
{
\begin{align}\label{eq:output_mask}
	\hat{P} = \hat{X} \otimes \tilde{P}
\end{align}
}%
where $\hat{X} \in \mathbb{R}^{n \times m}$, $\tilde{P} \in \mathbb{R}^{m \times H \times W}$, $\hat{P} \in \mathbb{R}^{n \times H \times W}$, and $\otimes$ indicates the tensor contraction operator along the last and the first dimensions of $\hat{X}$ and $\tilde{P}$, respectively.
Here $H$ and $W$ denote the height and width of the input image.
Each spatial slice of $\hat{P}$ denotes the matched mask corresponding to a particular template.
This process is shown in the yellow box of Fig.~\ref{fig:model}.
The matched mask for the template will be used to compute the IoU score, shown in Eq.~(\ref{eq:cost_matrix}), at the next time step. 
Therefore, we propagate the latest mask information over time.


Given the output mask from matching, we then refine it using the template of the same instance.
In particular, we construct the input by stacking multi-scale image features from the backbone, the matched mask and template mask. The multi-scale features are extracted from the last layer of the conv 2-5 blocks in the feature extractor backbone, respectively.   
Inspired by RVOS, we adopt a decoder containing four ConvLSTM~\cite{NIPS2015_5955} layers as the refinement module to predict the masks of all objects at each time step and carry over the memory and hidden states to the next time step. 



\section{Experiments}

In this section, we compare our {\ourmodelshort} with a wide range of recent competitors on YouTube-VOS, DAVIS 2017 and SegTrack v2 datasets. 
YouTube-VOS has $3,471$ and $474$ videos for training and validation, respectively. Among the $91$ object categories in the validation set, $65$ are seen in the training set while the other $26$ are unseen. 
DAVIS 2017 has $60$ and $30$ video sequences for training and validation respectively and the average video length is around $70$.
The average number of instances per sequence are $2.30$ and $1.97$ for training and validation, respectively.
For the SegTrack v2 dataset, there are $14$ low resolution videos ($947$ frames in total) with $24$ generic foreground objects.
All of our experiments are conducted on NVIDIA Titan XP GPUs.

\begin{table}[t]
\resizebox{\linewidth}{!}{
\begin{tabular}{@{}l|c|cccc|c|c@{}}
\hline
\toprule
& \texttt{OL} & $\mathcal{J}_{\mathcal{S}}$ & $\mathcal{J}_{\mathcal{U}}$ & $\mathcal{F}_{\mathcal{S}}$ & $\mathcal{F}_{\mathcal{U}}$  & $\mathcal{G}_{\mathcal{M}}$ & \texttt{FPS} \\ 
\midrule
\midrule
OSMN~\cite{Yang2018EfficientVO}  & \xmark  & 60.0 & 40.6 & 60.1 & 44.0 & 51.2 & 8.0 \\
SiamMask~\cite{wang2019fast} & \xmark  & 60.2 & 45.1 & 58.2 & 47.7& 52.8 & \textbf{55} \\
RGMP~\cite{wug2018fast} & \xmark  & 59.5&-&45.2&-&53.8& 7  \\
OnAVOS~\cite{voigtlaender2017online} & \cmark & 60.1 & 46.6 & 62.7 & 51.4 & 55.2 & - \\
RVOS~\cite{Ventura_2019_CVPR} & \xmark&63.6&45.5&67.2&51.0&56.8 &24  \\ 
S2S~\cite{Xu2018YouTubeVOSSV} & \xmark  & 66.7&48.2&65.5&50.3&57.7&6  \\
OSVOS~\cite{caelles2017one} & \cmark & 59.8 & 54.2 & 60.5 & 60.7 & 58.8 & - \\
S2S~\cite{Xu2018YouTubeVOSSV} & \cmark  & \textbf{71.0}&\textbf{55.5} &\textbf{70.0}&\textbf{61.2}&\textbf{64.4}& - \\ 
DMM-Net & \cmark  & 59.2 & 47.6 & 62.6 & 53.9 & 55.8&  -  \\
DMM-Net+ & \xmark  & 58.3 & 41.6 & 60.7 & 46.3 & 51.7 &  12  \\
DMM-Net+ & \cmark  & 60.3 & 50.6 &	63.5 &	57.4  & 58.0&  -  \\
\bottomrule
\end{tabular}
}
\caption{Results on YouTube-VOS (validation set) and frame-per-second (FPS) during inference for methods withour online learning. `$\mathcal{S}$' and `$\mathcal{U}$' denote the seen and unseen categories. `OL': online learning. `+' means we use ResNet-101 as feature extractor instead of ResNet-50}
\vspace{-5mm}
\label{tab:youtubevos}
\end{table}

\begin{table}[t]
\centering
{%
\begin{tabular}{@{}c|c|c|c @{}}
\hline
\toprule
Methods & $\mathcal{J}_{\mathcal{M}}$ & $\mathcal{F}_{\mathcal{M}}$ & $\mathcal{G}_{\mathcal{M}}$ \\ 
\midrule
\midrule
MaskRNN~\cite{hu2017maskrnn} &  45.5 & - & - \\ 
OSMN~\cite{Yang2018EfficientVO} &  52.5 & 57.1 & 54.8 \\
FAVOS~\cite{cheng2018fast} &  54.6 & 61.8 & 58.2  \\
VideoMatch~\cite{hu2018videomatch}  & 56.5 & - & - \\
MSK~\cite{perazzi2017learning}  & 63.3 & 67.2 & 65.3 \\
RGMP~\cite{wug2018fast} & 64.8 & 68.6 & 66.7\\
FEELVOS~\cite{feelvos2019}$^{\ast}$ & 65.9  & 72.3  & 69.1 \\
DyeNet~\cite{li2018video} & 67.3 & 71.0 & 69.1 \\
\ourmodelshort & \textbf{68.1} & \textbf{73.3} & \textbf{70.7}\\
\bottomrule
\end{tabular}
}
\caption{Results without online learning on the validation set of DAVIS 2017 dataset. FEELVOS$^{\ast}$ also reports another better performed model which is trained on YouTube-VOS~\cite{xu2018youtube}. `-' means no public results available.}
\vspace{-0.2cm}
\label{table:DAVIS}
\end{table}


\subsection{Implementation Details}~\label{sect:implement}

We first introduce implementation details of our model.

\vspace{-2mm}
\paragraph{Mask Proposal Generation}
In the stage of mask proposal generation, we use Mask R-CNN with ResNeXt-101-FPN as the backbone which is pretrained on COCO dataset~\cite{lin2014microsoft}.
Score threshold of the ROI head is set to $0$.
We resize the input image such that its short-side is no larger than $800$. 
We first train the class-agnostic binary mask proposal network on COCO.
Following the strategy used in~\cite{wang2019fast}, we then finetune the proposal network on the combination of COCO and YouTube-VOS with learning rate $0.02$, batch size $8$ and number of training iteration $200,000$.

\paragraph{Differentiable Mask Matching}
For the feature extractor $f_{\theta}$, we use a COCO-pretrained Mask R-CNN with a ResNet-50-FPN backbone. 
We also try a ResNet-101 backbone of which the weights are initialized from the released model of RVOS~\cite{Ventura_2019_CVPR}.
We denote this model as DMM-Net+. 
Note that it is possible to share the backbone between the proposal network and the feature extractor of matching such that the overall model is more compact. 
However, sharing backbone leads to worse results in our experiments which may suggest that generating good proposals and learning good feature for matching require different representations.
After we obtain the proposals for each frame, we perform ROI pooling for each proposal to extract multi-scale feature 
from the backbone and then average the feature spatially to obtain a single feature vector. Similar to the feature fed to the refinement layer, we obtain the proposals feature from the last layer of conv2-5 block in the backbone. 
Input image of the feature extractor is resized such that the short-side is no larger than $480$ and $800$ for DAVIS 2017 and SegTrack v2, respectively. 
For YouTube-VOS, we resize image to $255\times448$ in order to have a fair comparison with S2S~\cite{Xu2018YouTubeVOSSV} and RVOS~\cite{Ventura_2019_CVPR}.
The score weight $\lambda$ used to compute the matching cost in Eq.~(\ref{eq:cost_matrix}) is set to $0.3$ for DAVIS 2017 and YouTube-VOS, and $0.9$ for SegTrack v2.
For the mask matching, we set $N_{\text{grad}} = 40$, $N_{\text{proj}} = 5$ and learning rate $\alpha=0.1$. Ablation study on the hyperparameters of the matching is left in the supplementary material. 
After the matching assignment matrix $\hat{X}$ is obtained, we also found it helpful to remove the non-confident matching by applying a differentiable masking  $\hat{X} = \hat{X} \cdot \mathbf{1} [\hat{X} = \text{max}(\hat{X})]$.

\begin{table}[t]
\centering
{%
\begin{tabular}{@{}c|c|c|c@{}}
\hline
\toprule
\multirow{2}{*}{Methods} & Online & \multirow{2}{*}{mIoU$^{\ast}$} & \multirow{2}{*}{mIoU$^{\dagger}$} \\
 & Learning & & \\ 
\midrule
\midrule
OSVOS~\cite{caelles2017one} & \cmark &61.9 & 65.4 \\
OFL~\cite{tsai2016video} & \cmark & 67.5 & - \\
MSK~\cite{perazzi2017learning} & \cmark & 67.4 &70.3 \\
RGMP~\cite{wug2018fast} & & - &71.1 \\
MaskRNN~\cite{hu2017maskrnn} &  & 72.1 & - \\
LucidTracker~\cite{khoreva2017lucid} & \cmark & - & 77.6 \\
DyeNet~\cite{li2018video} &  &- &78.3 \\
DyeNet~\cite{li2018video} & \cmark & - & \textbf{78.7} \\
\ourmodelshort &  & \textbf{76.8} & 76.7 \\
\bottomrule
\end{tabular}
}
\caption{Results on the SegTrack v2 dataset. mIoU$^{\ast}$ is averaged over all frames whereas mIoU$^{\dagger}$ is averaged over all instances.}
\vspace{-0.5cm}
\label{table:segtrack}
\end{table}


\paragraph{Refinement Network}\label{sec:refine}
A light version of refinement network containing only four ConvLSTM layers is used for experiments on YouTube-VOS to reduce the computational cost. 
The weights are randomly initialized and trained with the matching layer in an end-to-end manner. 
Input images are resized to be $255 \times 448$ for both training and inference. 
The refinement network outputs our final predicted masks. 

A heavier version of refinement network is used for model trained on DAVIS 2017. We follow the U-Net style architecture as RGMP~\cite{wug2018fast} and initialize the weights from their released model. The refinement network is also trained together with the matching layer. 

\paragraph{Fine-tune}\label{para:ft}
Since our {\ourmodelshort} is end-to-end differentiable, we fine-tune it on the training sets of YouTube-VOS and DAVIS 2017.
We use the Adam~\cite{kingma2014adam} optimizer with the learning rate $1.0e^{-4}$ for the weights initialized from pre-trained model and $1.0e^{-3}$ for the weights from random initialization. 
We set batch-size to $24$ and train on YouTube-VOS dataset for $10$ epochs in total. 
Data augmentation such as random affine transformation is applied during training. 
On DAVIS 2017, we use the same optimizer with learning rate $1.0e^{-7}$ and batch size $1$. We fine-tune the model for $8$ epochs in total.

\paragraph{Online Learning}
For online learning, we train both the proposal generator and the DMM-Net with the refinement module on the first frame of the validation set. We use the same learning rate and batch size as  Section~\ref{para:ft} except that the number of epochs is $100$ for online learning. 

\paragraph{Evaluation}
For YouTube-VOS and DAVIS 2017, we follow~\cite{pont2017} and use the region ($\mathcal{J}$), boundary ($\mathcal{F}$) and their average ($\mathcal{G}$) score as the metrics.
For SegTrack v2, there are two types of mean IoU adopted in the existing literature. 
Specifically, one can compute the IoU averaging over all instances per frame and then average over all frames as in~\cite{hu2017maskrnn}, denoted as mIoU$^{\ast}$.
One can also compute the IoU per instance, average over the frames where the instance has appeared, and then average over all instances as in~\cite{li2018video}, denoted as mIoU$^{\dagger}$. 
We report both metrics on this dataset.

\subsection{Main Results}


\begin{figure*}[t]
	\centering
	\begin{tabular}{@{\hspace{0mm}}c@{\hspace{0 mm}}c@{\hspace{0 mm}}c@{\hspace{0mm}}c@{\hspace{0mm}}c}
		\includegraphics[width=0.195\linewidth]{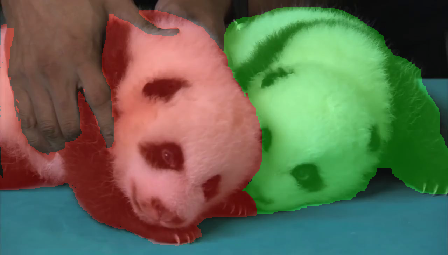}&
		\includegraphics[width=0.195\linewidth]{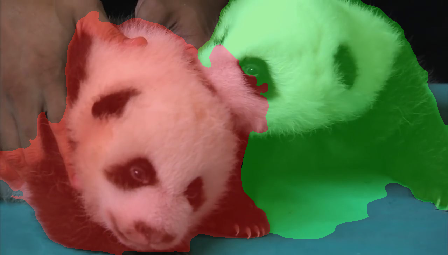}&
		\includegraphics[width=0.195\linewidth]{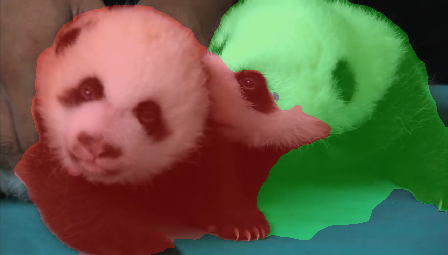}&
		\includegraphics[width=0.195\linewidth]{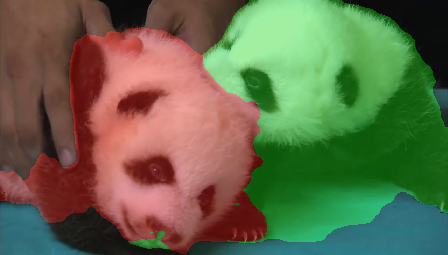}&
		\includegraphics[width=0.195\linewidth]{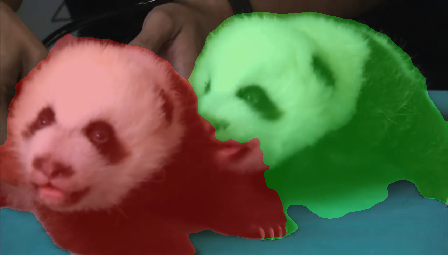}\\[-1.0mm]
		\includegraphics[width=0.195\linewidth]{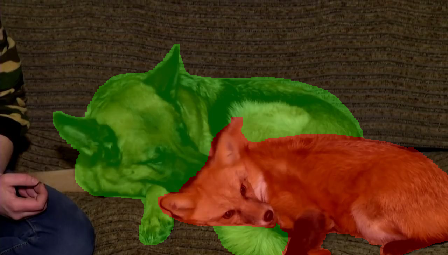}&
		\includegraphics[width=0.195\linewidth]{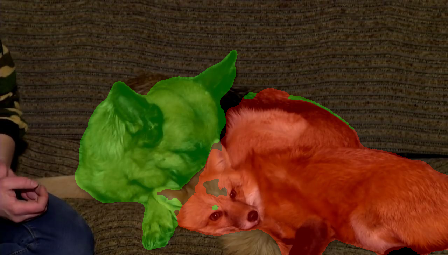}&
		\includegraphics[width=0.195\linewidth]{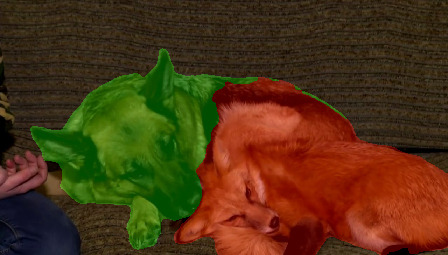}&
		\includegraphics[width=0.195\linewidth]{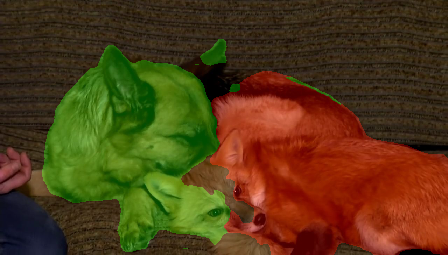}&
		\includegraphics[width=0.195\linewidth]{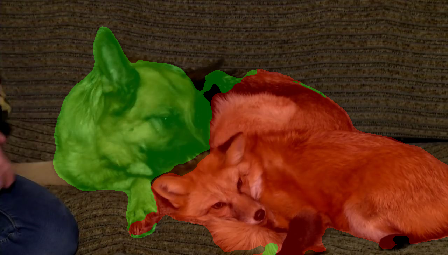}\\[-1.0mm]
		\includegraphics[width=0.195\linewidth]{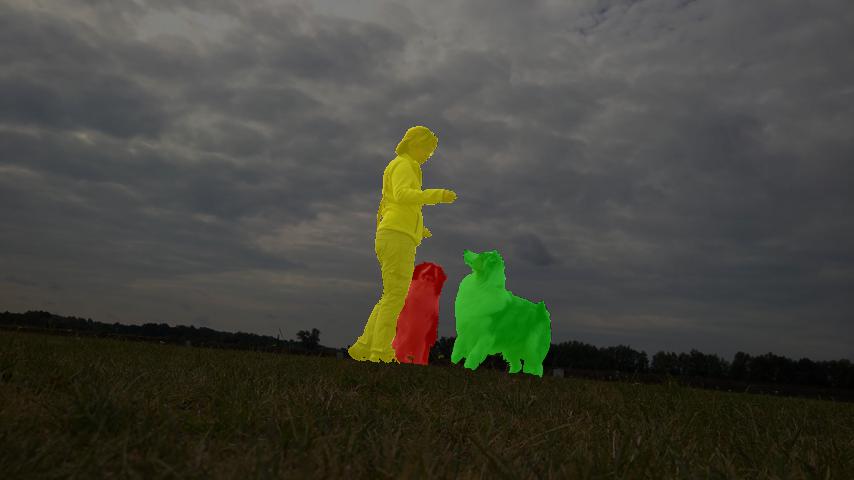}&
		\includegraphics[width=0.195\linewidth]{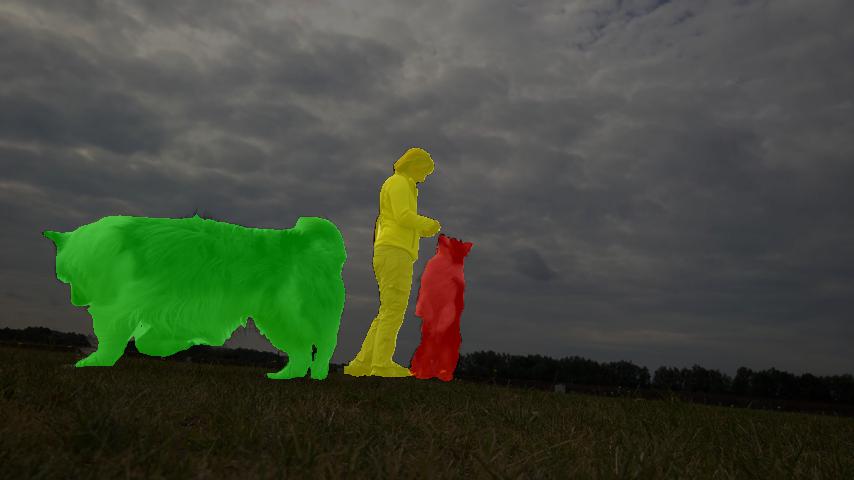}&
		\includegraphics[width=0.195\linewidth]{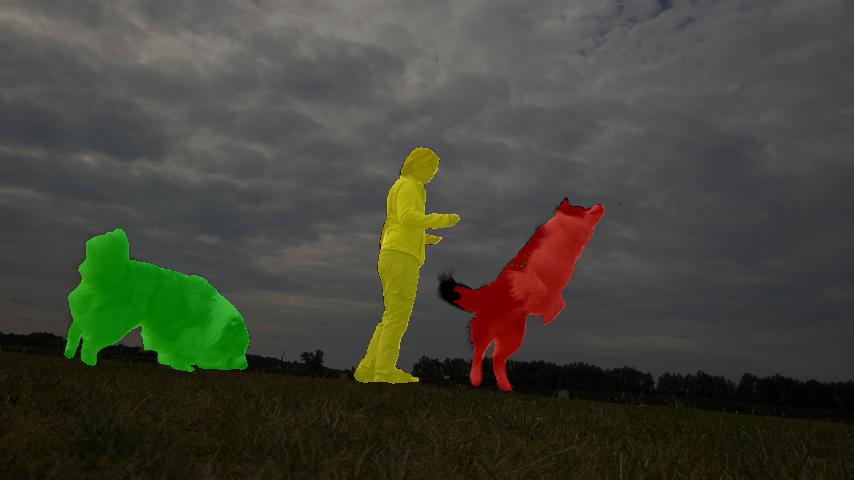}&
		\includegraphics[width=0.195\linewidth]{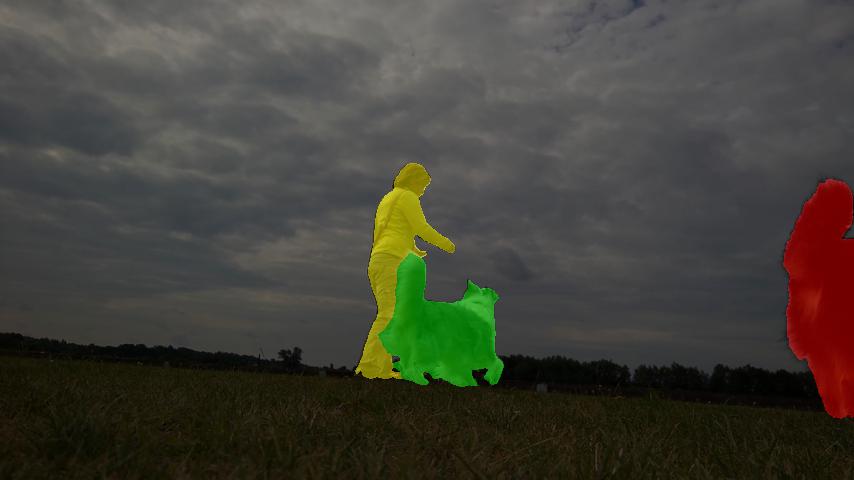}&
		\includegraphics[width=0.195\linewidth]{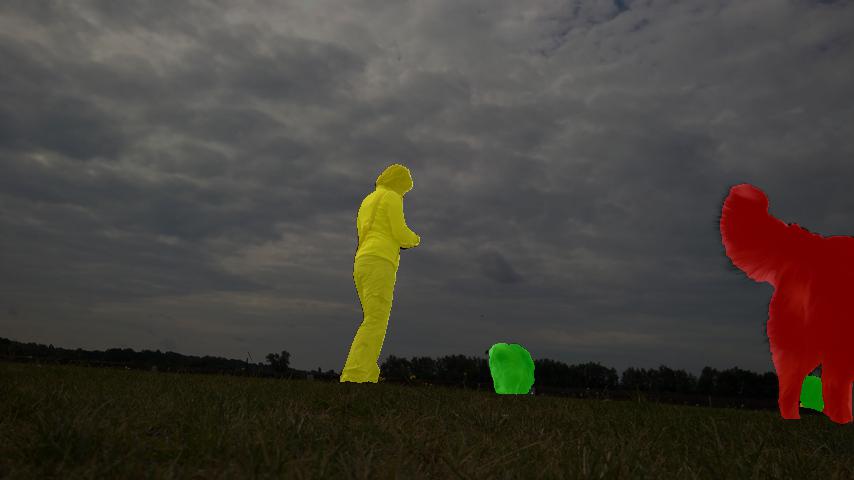}\\[-1.0mm]
		\includegraphics[width=0.195\linewidth]{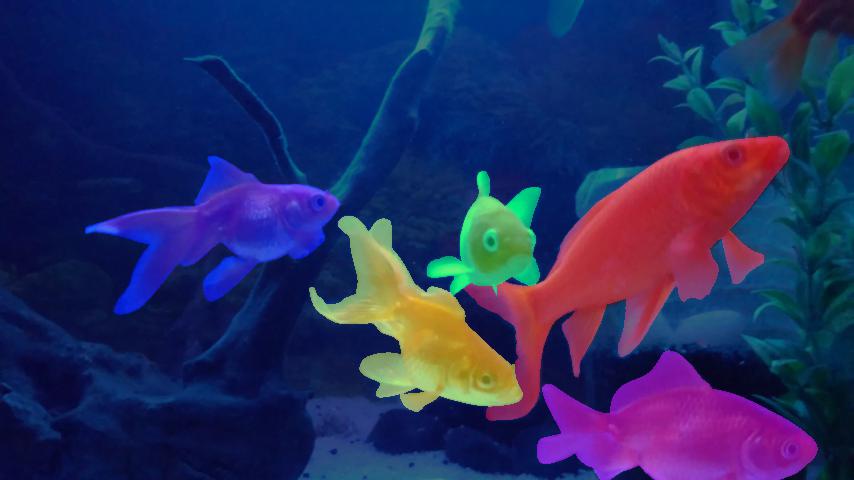}&
		\includegraphics[width=0.195\linewidth]{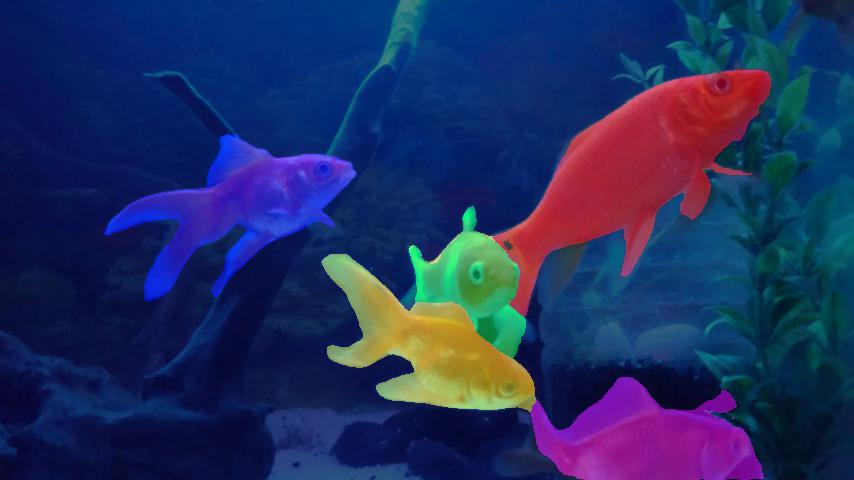}&
		\includegraphics[width=0.195\linewidth]{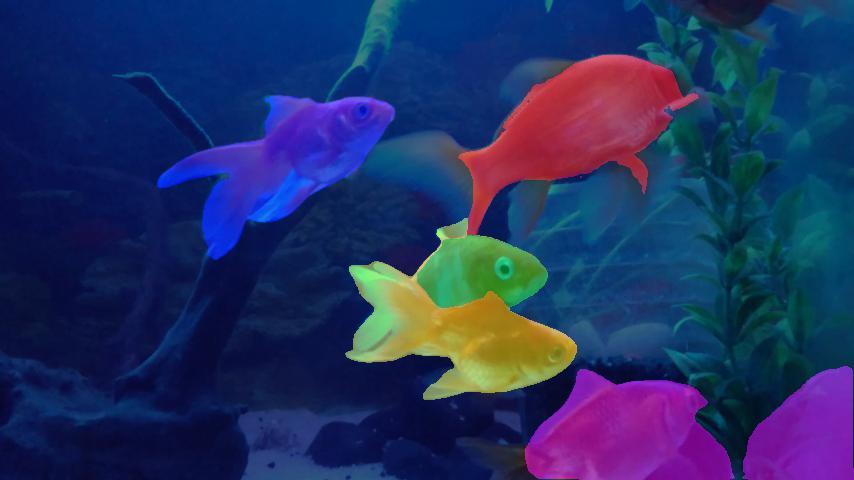}&
		\includegraphics[width=0.195\linewidth]{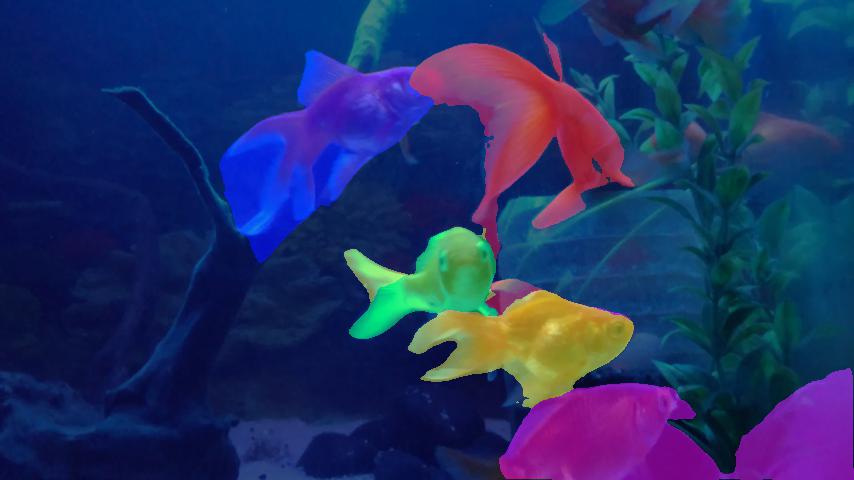}&
		\includegraphics[width=0.195\linewidth]{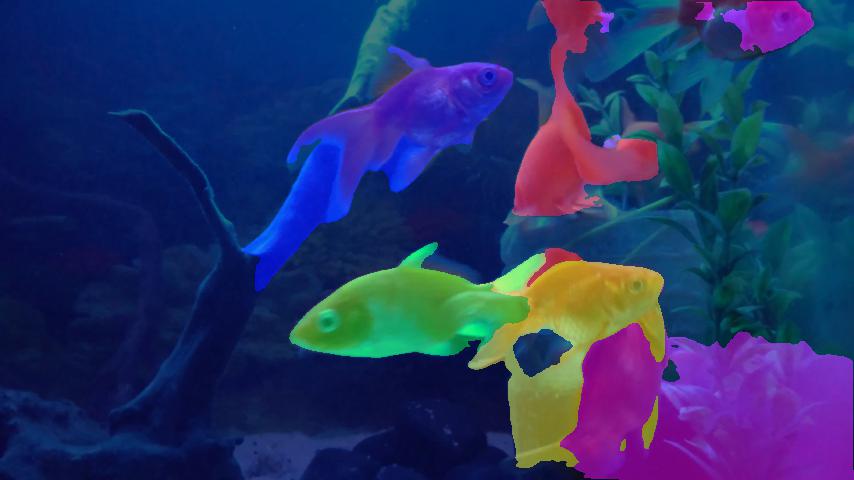}\\[-1.0mm]
		\includegraphics[width=0.195\linewidth]{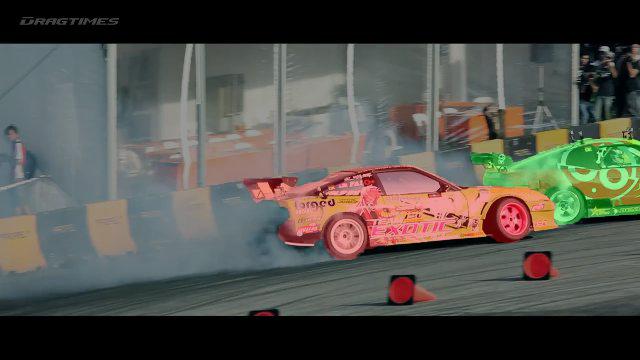}&
		\includegraphics[width=0.195\linewidth]{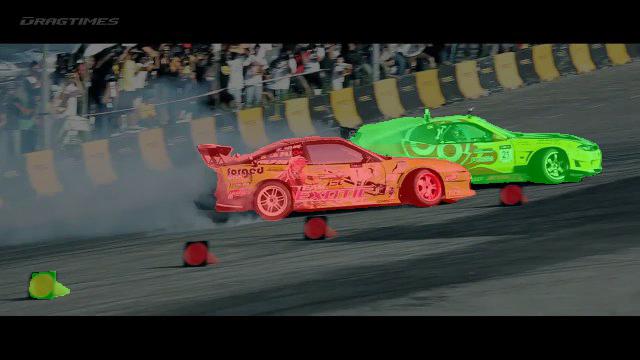}&
		\includegraphics[width=0.195\linewidth]{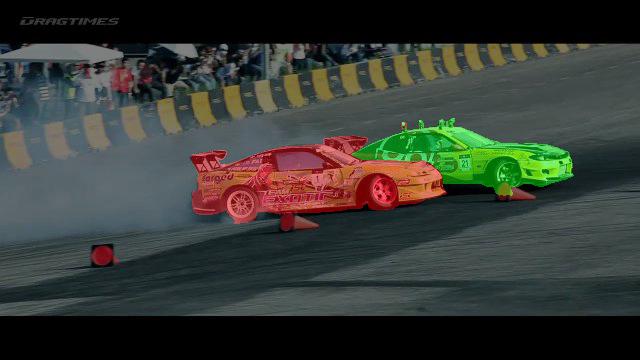}&
		\includegraphics[width=0.195\linewidth]{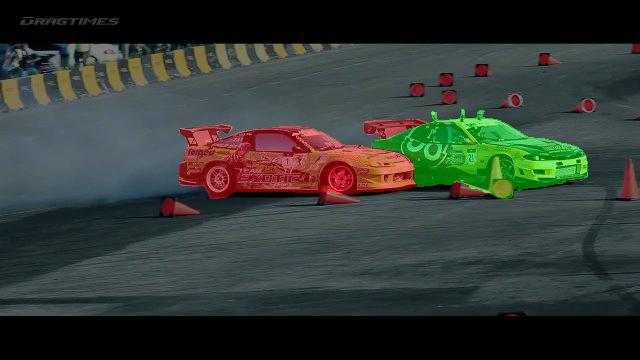}&
		\includegraphics[width=0.195\linewidth]{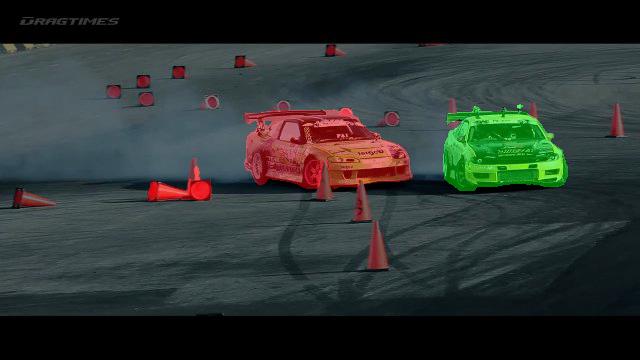}\\[-1.0mm]
		\includegraphics[width=0.195\linewidth]{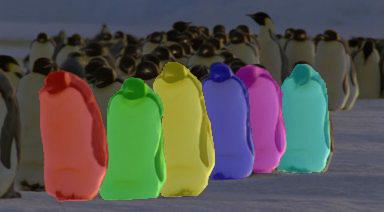}&
		\includegraphics[width=0.195\linewidth]{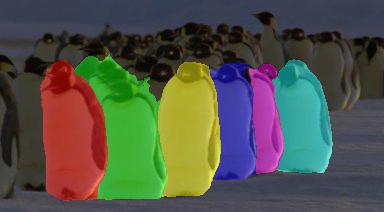}&
		\includegraphics[width=0.195\linewidth]{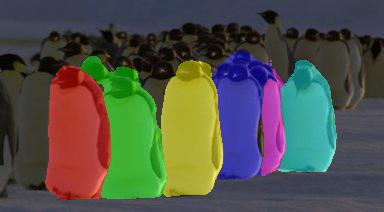}&
		\includegraphics[width=0.195\linewidth]{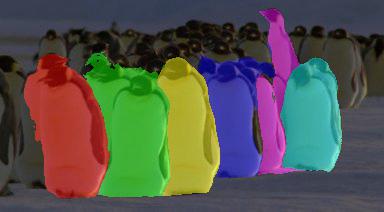}&
		\includegraphics[width=0.195\linewidth]{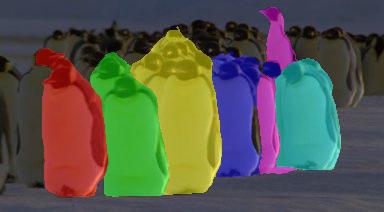}\\
		{\footnotesize (a) 0\% } &{\footnotesize (b) 25\%} &{\footnotesize (c) 50\%}&{\footnotesize (d) 75\%}& {\footnotesize (e) 100\% } \\[0.7mm]
	\end{tabular}
	\caption{Visualization of our results on YouTube-VOS, DAVIS 2017 and SegTrack v2 at different time steps (percentage \wrt the whole video length). The first $2$, the middle $2$ and the last $2$ rows correspond to the YouTube-VOS, DAVIS 2017 and SegTrack v2 datasets respectively.} 
	\vspace{-0.5cm}
	\label{fig:vis}
\end{figure*}

\paragraph{YouTube-VOS}
We fine-tune both our DMM-Net and proposal net on YouTube-VOS. We first split the official training set into train-train, train-val and train-test splits. Our train-val split consists of 200 videos while the train-train split consists of 3000 videos. Our training is performed on the train-train split, and the best model is selected based on the performance on the train-val split. The final performance on the official validation set is reported in Table~\ref{tab:youtubevos}. 
Compared to the state-of-the-art method S2S, our model achieves competitive segmentation metrics with double speed.
In general, we obtain a good trade off between the performance and running time. 
Moreover, we found that using a stronger backbone, \ie, DMM-Net+, can further boost the performance.

\paragraph{DAVIS 2017}
We compare with a wide range of recent competitors on the validation set of DAVIS 2017. For experiments on DAVIS, we only fine-tune our DMM-Net on the training set of DAVIS 2017 and use the proposal generator pretrained on COCO. 
The models without online learning are listed in Table~\ref{table:DAVIS}. 
From the table, it is clear that without online learning, our method achieves the state-of-the-art performance.
In Fig.~\ref{fig:vis}, we show the qualitative results of our {\ourmodelshort} at different time steps (uniformly sampled percentage \wrt the whole video length) of each video sequence.
From the figure, we can see that our model consistently keeps a very good segmentation quality as time goes on.
We also show a visual comparison at the last frame of the \textit{soapbox} sequence with other strong competitors in Fig.~\ref{fig:vis_comp}.
It is clear that our model does a better job in segmenting the details of the persons and the soapbox.
However, some failure cases still exist. 
For example, in the $100\%$ column and $4$th row of Fig.~\ref{fig:vis}, the segmentation of the goldfish in the bottom-right corner is unsatisfying.

\paragraph{SegTrack v2}
We test our {\ourmodelshort} model (fine-tuned on DAVIS 2017 training set) directly on the full SegTrack v2 dataset.
We do not perform any fine-tuning on this dataset. 
Moreover, for simplicity, we again do not adopt any online learning such that we could fully test the generalization ability of our model.
The quantitative results are listed in Table~\ref{table:segtrack}.
From the table, we can see that without any fine-tuning and online learning, our {\ourmodelshort} achieves comparable performance to the state-of-the-art methods.
We show some visual examples in the bottom two rows of Fig.~\ref{fig:vis}.
We can see that our model again has a consistently high segmentation quality across different time steps.

\subsection{Ablation Study}\label{sec:ablation}
In this section, we conduct thorough ablation study to justify the design choice and hyperparameters of our model.
\newcommand{\colwidthA}{0.3cm}
\newcommand{\colwidthB}{0.7cm}
\begin{table}[t]
\centering
\resizebox{\linewidth}{!}{
\begin{tabular}{c|cc|cc|cc}
\hline \toprule
\multicolumn{1}{c|}{Matching} & \multicolumn{2}{c|}{DMM-Net} & \multicolumn{2}{c|}{Prop. Net} & \multicolumn{2}{c}{Train-val} \\[.3em]    
\cline{2-7} 
 & \texttt{Ft.} & \texttt{Unroll} & \texttt{Arch.} & \texttt{+ytb} & $\mathcal{J_M}$ & $\mathcal{F_M}$ \\ 
\midrule \midrule  
Greedy &      - & - & X101&  & 57.1 & 68.1 \\
Hungarian &      - & - & X101&  & 57.3 & 68.4 \\
Ours & \xmark & - & X101&  & 57.3 & 68.3 \\
Ours& \cmark & 2 & R50 &  & 58.5  & 71.4 \\
Ours& \cmark & 2 & X101&     & 59.0 & 71.7 \\
Ours& \cmark & 3 & X101&  & 58.2  & 71.4 \\
Ours& \cmark & 2 & X101& \cmark & \textbf{60.2}  & \textbf{73.0} \\
 \bottomrule
\end{tabular}
}
\caption{Ablation study evaluated on our train-val split of YouTube-VOS. Prop. Net: mask proposal network. `+ytb': using YouTube-VOS train-train split during the training of proposal net or not. `Ft.': fine-tuning, `Arch.': architecture for the proposal net, `R50': ResNet-50, `X101': ResNetXt-101.}
\vspace{-0.2cm}
\label{table:DAVIS_ablation_1}
\end{table}

\paragraph{Greedy vs. Hungarian vs. Our Matching Layer}
We first test the matching layer against the optimal matching using the Hungarian method and the popular greedy approximation during inference.
For a fair comparison, we use the same set of mask proposals, the same feature extractor network pretrained on COCO dataset.
We show the mean of $\mathcal{J}$ and $\mathcal{F}$ scores on YouTube-VOS dataset in Table~\ref{table:DAVIS_ablation_1}. All the ablation results are obtained by training on our train-train split and evaluating on train-val split. 
From the table, we can see that our matching layer has similar performance compared to the optimal matching and is superior to the greedy one during inference.

\paragraph{End-to-End Fine-tuning}
We now study the effect of fine-tuning the whole model on the train-train split of YouTube-VOS.
As shown in Table~\ref{table:DAVIS_ablation_1}, the performance is improved significantly, which verifies the benefits of the end-to-end training and the differentiability of our matching layer.

\paragraph{Mask Proposal Network}
We also try different backbones for the mask proposal network and fine-tune the network on YouTube-VOS dataset. As shown in Table~\ref{table:DAVIS_ablation_1}, the performance gain is pretty high by fine-tuning the proposal net on large scale video dataset such as YouTube-VOS.

\paragraph{Number of Unrolled Steps} 
At last, we investigate the number of unrolled steps of the refinement network during training.
As shown in Table~\ref{table:DAVIS_ablation_1}, unrolling more than $2$ step seems not helpful and increases the memory cost significantly.
Note that during test we unroll from the beginning to the end of the video sequence.







\section{Conclusion}
In this paper, we propose the {\ourmodel} ({\ourmodelshort}) for solving the problem of video object segmentation.
Relying on the pre-computed masks proposals, {\ourmodelshort} first conducts the mask matching between proposals and templates via a projected gradient descent method which is guaranteed to converge and fully differentiable.
It enables the learning of the cost matrix of matching.
Based on the template mask, we then refine the current matched mask to further improve the segmentation quality.
We demonstrate that our model achieves the state-of-the-art or comparable performances under different settings of several challenging benchmarks.
In the future, we would like to apply our differentiable matching layer to other backbone networks for the purpose of further boosting the performance.
Moreover, exploring the mask matching in a longer temporal window, \ie, a multi-partite matching problem similar to tracking, would be very interesting.

\subsubsection*{Acknowledgments}
We gratefully acknowledge support from Vector Institute.
RL was supported by Connaught International Scholarships, RBC Fellowship and NSERC. 
SF acknowledges the Canada CIFAR AI Chair award at Vector Institute. Part of this work was also supported by NSERC Cohesa grant, and Samsung. 
We are grateful to Relu Patrascu for infrastructure support.

\clearpage

\bibliographystyle{ieee_fullname}
\bibliography{egbib}

\section{Proof of Theorem 1}

In this section, we prove our main result, \ie, Theorem~\ref{them:converge}, based on the established convergence results of \textit{Dykstra's algorithm} and the inexact projected gradient method for the general constrained convex minimization~\cite{patrascu2018convergence}.

\subsection{Convergence of Dykstra's Algorithm}
First, we state the convergence result of \textit{Dykstra's algorithm} from~\cite{deutsch1994rate} without proof as Lemma~\ref{lemma:dykstra} below.

\begin{lemma}\label{lemma:dykstra}
Suppose $K$ is the intersection of a finite number of closed half-spaces in a Hilbert space $X$. Starting with any point $x \in X$, there exists some constants $\rho > 0$ and $0 \le c < 1$, such that the sequence of iterates $\{x_{n}\}$ generated by Dykstra's algorithm satisfied the inequality,
\begin{align}
    \Vert x_{n} - \mathcal{P}_{K}(x) \Vert \le \rho c^{n}
\end{align}
for all $n$, where $\mathcal{P}_{K}(x)$ is the nearest point in $K$ to $x$.
\end{lemma}

\subsection{Convergence of Inexact Projected Gradient Method}
Now we turn to the inexact projected gradient method as described in~\cite{patrascu2018convergence} which aims at solving the following general constrained convex minimization,
\begin{align}
    \text{min}_{x \in \mathbb{R}^{n}} \quad f(x) \qquad \text{s.t.} \quad x \in \mathcal{C},
\end{align}
where $f:\mathbb{R}^{n} \rightarrow \mathbb{R}$ is convex and $\mathcal{C}$ is a general closed convex set.
We only state the results which are relevant to our case, \ie, $f$ is differentiable and has Lipschitz continuous gradient with constant $L_f$. 
In particular, we have,
\begin{align}\label{eq:Lipschitz}
    \Vert \nabla f(x) - \nabla f(y) \Vert  \le L_f \Vert x - y \Vert \qquad \forall x,y \in \mathbb{R}^{n} 
\end{align}
The inexact projected gradient method is stated as below.
\begin{algorithm}
\caption{: Inexact Projected Gradient Method for General Constrained Convex Minimization}
\begin{algorithmic}[1]
\STATE \textbf{Input:} $\delta > 0, N, x_{0}$
\STATE \textbf{For} $k = 1, 2, \dots, N-1$:
\STATE \qquad $z_{k} = x_{k-1} - \frac{1}{L_f} \nabla f(x_{k-1})$
\STATE \qquad Compute $x_{k}$ such that $\Vert x_{k} - \mathcal{P}_{\mathcal{C}}(z_{k}) \Vert \le \delta$
 \medskip
 \medskip
\STATE \textbf{Return} $\frac{1}{N} \sum_{k=0}^{N-1} x_{k}$
\end{algorithmic}
\end{algorithm}
Note that each projection step is inexact as it only requires $\delta$-approximation.
Therefore, our proposed Alg.~\ref{alg:matching} can be regarded as a special instance of this general algorithmic framework of projected gradient method.
We state the convergence result of this method from~\cite{patrascu2018convergence} without proof as Lemma~\ref{lemma:inexact} below.

\begin{lemma}\label{lemma:inexact}
Assuming the objective $f$ is differentiable and has Lipschitz continuous gradient with constant $L_f$. Let $\{x_{k}\}$ be the sequence generated by the inexact projected gradient method. Given inner accuracy $\delta > 0$ and any $k \ge 0$, assuming $\Vert \nabla f(x^{\ast}) \Vert \le L_{f}r_{0}$, then the following sublinear estimates for feasibility and suboptimality in $\hat{x}_{k} = \frac{1}{k} \sum_{i=0}^{k} x_{i}$ hold:
\begin{align}
\Vert \hat{x}_{k} - \mathcal{P}_{\mathcal{C}}(\hat{x}_{k}) \Vert & \le \delta, \nonumber \\
f(\hat{x}_{k}) - f(x^{\ast}) & \ge - \Vert \nabla f(x^{\ast}) \Vert \delta, \nonumber \\
f(\hat{x}_{k}) - f(x^{\ast}) & \le \frac{2L_{f}r_{0}^2}{k} + 5L_{f}r_{0}\delta + 5kL_{f}\delta^2,
\end{align}
where $r_0 = \Vert x_{0} - x^{\ast} \Vert$.
\end{lemma}
Note that the sublinear convergence rate is obtained provided that $\delta = O(\frac{1}{k})$.

\subsection{Our Result}

Now we are ready to prove our main result.
Recall that our objective is $f(X) = \Tr(CX^{\top})$.
The gradient is $\nabla f(X) = C$.
Therefore, any $L_f \ge 0$ is a Lipschitz constant satisfying Eq.~(\ref{eq:Lipschitz}).
Also, our constraint set $\mathcal{C}$ is a closed convex set.

\begin{theorem}
Let $r_{0} = \Vert X^{0} - X^{\ast} \Vert_{F}$ where $X^{0}$ and $X^{\ast}$ are the initial and optimal assignment matrices respectively.
Let the learning rate $0 < \alpha < \min (15 r_{0}, r_{0}/\Vert C \Vert_{F})$.
There exists some constants $0 \le c < 1$ and $\rho > 0$ such that the at any outerloop iteration $i$ of Alg.~\ref{alg:matching}, the error of projection $\Vert X^{i} - \mathcal{P}_{\mathcal{C}}(X^{i}) \Vert_{F} \le \delta = \rho c^{N_{\text{proj}}}$ where $X^{i}$ and $\mathcal{P}_{\mathcal{C}}(X^{i})$ are the assignment matrix and its correct projection onto $\mathcal{C}$ respectively.
Moreover, for any $0 < \epsilon < 1$, there exists a $N_{\text{proj}} \ge \log_{1/c} \left( \rho  \sqrt{\frac{15K}{\alpha \epsilon}} \right)$ such that,
\begin{align}
    \delta \le \frac{\alpha \epsilon }{15 r_{0}} \nonumber
\end{align}
and after at most $K$ iterations where
\begin{align}
    K = \left\lceil {\frac{6r_{0}^{2}}{\alpha \epsilon}} \right\rceil, \nonumber
\end{align}
the output of Alg.~\ref{alg:matching} $\hat{X}$ is $\epsilon$-optimal, \ie, $\Vert \hat{X} - \mathcal{P}_{\mathcal{C}}(\hat{X}) \Vert_{F} \le \epsilon$ and $\vert \Tr{\left( C \hat{X}^{\top} \right)} - \Tr{\left( CX^{\ast\top} \right)} \vert \le \epsilon$.
\end{theorem}

\begin{proof}
At outer step $i$ of Alg.~\ref{alg:matching}, we run \textit{Dykstra's algorithm} for $N_{\text{proj}}$ inner steps.
We know from Lemma~\ref{lemma:dykstra} that there exists a $\rho_{i} > 0$ and $0 \le c_{i} < 1$ such that,
\begin{align}
    \Vert X^{i} - \mathcal{P}_{\mathcal{C}}(X^{i}) \Vert \le \rho_{i} c_{i}^{N_{\text{proj}}}
\end{align}
Denoting $\delta_{i} = \rho_{i} c_{i}^{N_{\text{proj}}}$, $\rho, c = \text{argmax}_{(\rho_{i}, c_{i}), i=1,\dots,N_{\text{grad}}} \delta_{i}$, and $\delta = \rho c^{N_{\text{proj}}}$, we have,
\begin{align}
    \max_{i = 1, \cdots, N_{\text{grad}}} \Vert X^{i} - \mathcal{P}_{\mathcal{C}}(X^{i}) \Vert_{F} \le \delta,
\end{align}
where $\rho > 0$ and $0 \le c < 1$.

Since $\delta$ is monotonically decreasing \wrt $N_{\text{proj}}$, there exists $N_{\text{proj}} \ge \log_{1/c} \left( \rho  \sqrt{\frac{15K}{\alpha \epsilon}} \right)$, for any $0 < \epsilon < 1$, such that,
\begin{align}\label{eq:delta}
    \delta \le \frac{\alpha \epsilon }{15 r_{0}}.
\end{align}

Our objective function $f(X) = \Tr(CX^{\top})$ is linear and constraint set $\mathcal{C}$ is closed and convex.
Therefore, our Alg.~\ref{alg:matching} is an instance of the inexact projected gradient method.
Moreover, since any $L_f \ge 0$ is a Lipschitz constant satisfying Eq.~(\ref{eq:Lipschitz}) in our case, we set $L_f = \frac{1}{\alpha}$ where $\alpha$ is is the learning rate and $0 < \alpha < \min (15 r_{0}, \frac{r_{0}}{\Vert C \Vert_{F}})$.
Now since all assumptions of Lemma~\ref{lemma:inexact} are satisfied, we apply it to our algorithm and obtain that,
\begin{align}\label{eq:main_result}
\Vert \hat{X} - \mathcal{P}_{\mathcal{C}}(\hat{X}) \Vert_{F} & \le \delta, \nonumber \\
f(\hat{X}) - f(X^{\ast}) & \ge - \Vert C \Vert_{F} \delta, \nonumber \\
f(\hat{X}) - f(X^{\ast}) & \le \frac{2r_{0}^2}{K\alpha} + 5\frac{r_{0}}{\alpha}\delta + 5\frac{K\delta^2}{\alpha},
\end{align}
where the $K$-step output of our Alg.~\ref{alg:matching} is $\hat{X} = \frac{1}{K} \sum_{i=0}^{K-1} X^{i}$ and we use the fact that $\nabla f(X^{\ast}) = C$.

If $K \ge \frac{6r_{0}^{2}}{\alpha \epsilon}$, then we have,
\begin{align}\label{eq:upper_bound_1}
    \frac{2r_{0}^2}{K\alpha} \le \frac{\epsilon }{3}. 
\end{align}
Due to Eq.~(\ref{eq:delta}), we have, 
\begin{align}\label{eq:upper_bound_2}
    5\frac{r_{0}}{\alpha}\delta \le \frac{\epsilon }{3}. 
\end{align}
Since $N_{\text{proj}} \ge \log_{1/c} \left( \rho  \sqrt{\frac{15K}{\alpha \epsilon}} \right)$, we have,
\begin{align}\label{eq:upper_bound_3}
    5\frac{K\delta^2}{\alpha} & = 5\frac{K}{\alpha} (\rho c^{N_{\text{proj}}})^{2} \nonumber \\
    & = 5\frac{K}{\alpha} \left( \sqrt{\frac{\alpha \epsilon}{15K}} \right)^{2} \nonumber \\
    & \le \frac{\epsilon }{3}. 
\end{align}
Therefore, with Eq.~(\ref{eq:upper_bound_1}), Eq.~(\ref{eq:upper_bound_2}) and Eq.~(\ref{eq:upper_bound_3}), we prove that $f(\hat{X}) - f(X^{\ast}) \le \epsilon$.

For the lower bound, from Eq.~(\ref{eq:main_result}) we have,
\begin{align}
f(\hat{X}) - f(X^{\ast}) & \ge - \Vert C \Vert_{F} \delta \nonumber \\
& \ge - \Vert C \Vert_{F} \frac{\alpha \epsilon }{15 r_{0}} \nonumber \\
& \ge - \frac{\epsilon }{15} \nonumber \\
& \ge - \epsilon, 
\end{align}
We prove the $\epsilon$-optimality \wrt the objective function.

Again, from Eq.~(\ref{eq:main_result}), we have,
\begin{align}
\Vert \hat{X} - \mathcal{P}_{\mathcal{C}}(\hat{X}) \Vert_{F} & \le \delta \nonumber \\
& \le \frac{\alpha \epsilon }{15 r_{0}} \nonumber \\
& \le \epsilon.
\end{align}
We prove the $\epsilon$-optimality \wrt the feasibility.

\end{proof}

Note that the constants (\eg, 6, 15) in Theorem~\ref{them:converge} do not matter that much as the inequality still holds by properly changing the constants in Lemma \ref{lemma:inexact} as discussed in \cite{patrascu2018convergence}.

\section{Additional Experiments}

\begin{figure*}[]
	\centering
	\begin{tabular}{@{\hspace{0mm}}c@{\hspace{0 mm}}c@{\hspace{0 mm}}c@{\hspace{0mm}}c@{\hspace{0mm}}c}
		\includegraphics[width=0.195\linewidth]{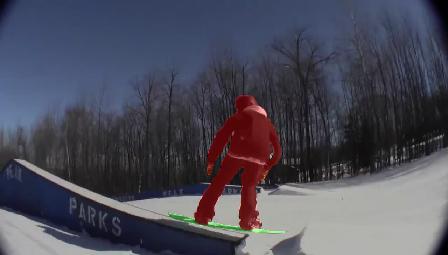}&
		\includegraphics[width=0.195\linewidth]{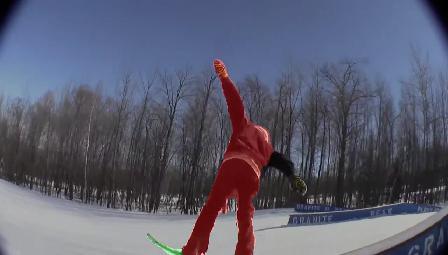}&
		\includegraphics[width=0.195\linewidth]{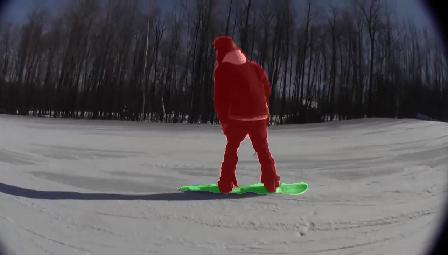}&
		\includegraphics[width=0.195\linewidth]{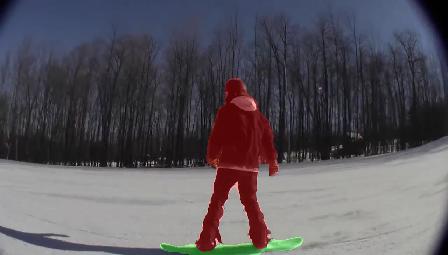}&
		\includegraphics[width=0.195\linewidth]{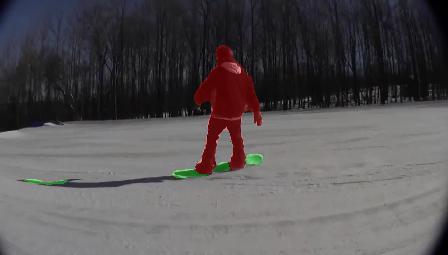}\\[-1.0mm]
		\includegraphics[width=0.195\linewidth]{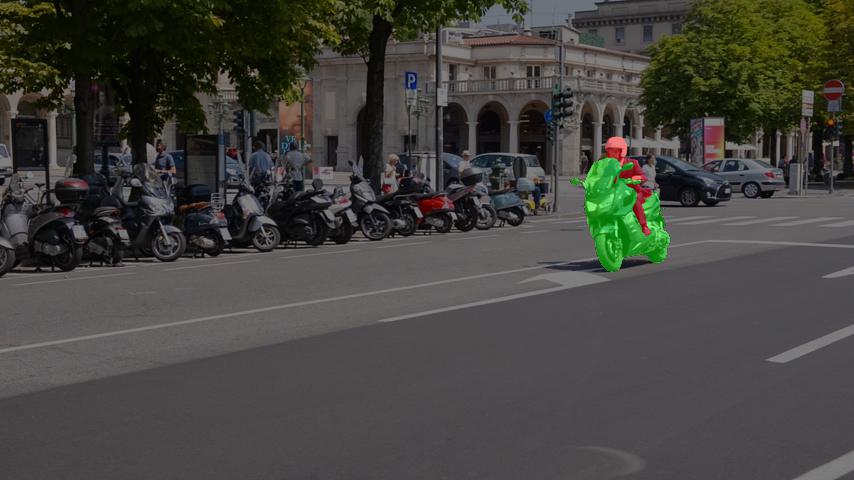}&
		\includegraphics[width=0.195\linewidth]{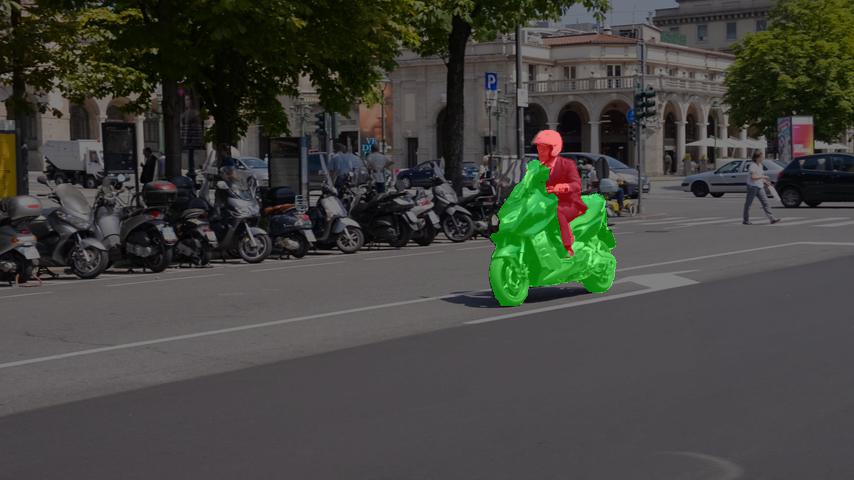}&
		\includegraphics[width=0.195\linewidth]{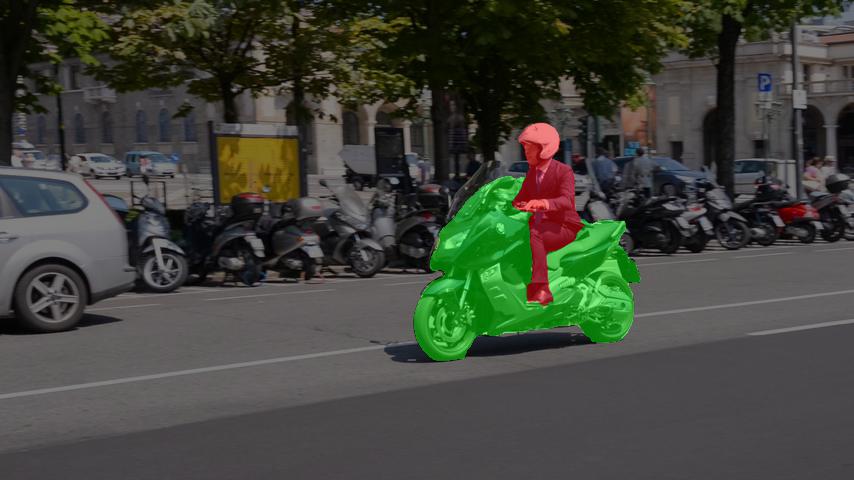}&
		\includegraphics[width=0.195\linewidth]{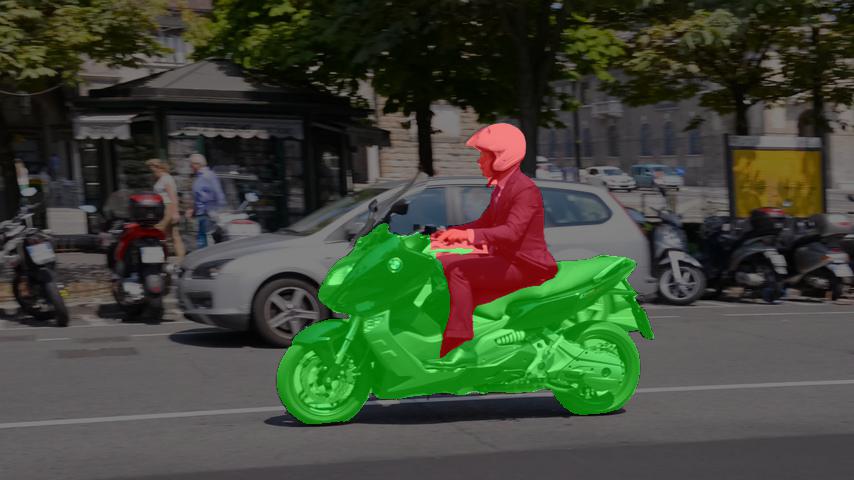}&
		\includegraphics[width=0.195\linewidth]{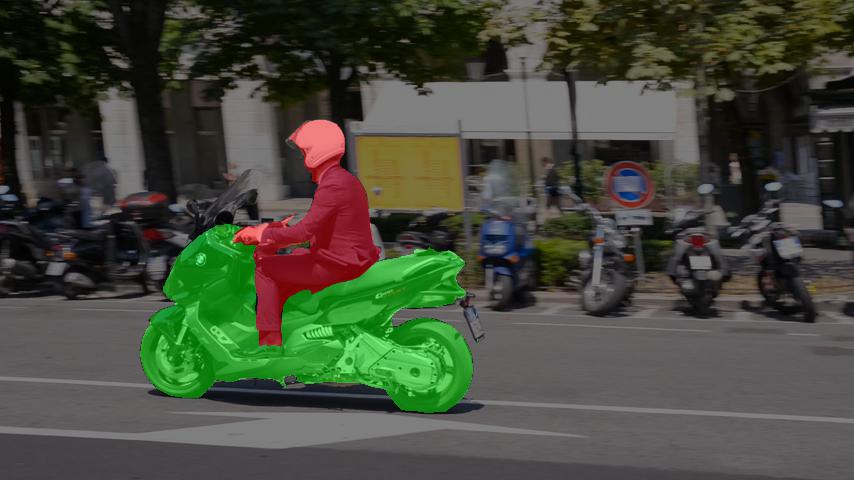}\\[-1.0mm]
		\includegraphics[width=0.195\linewidth]{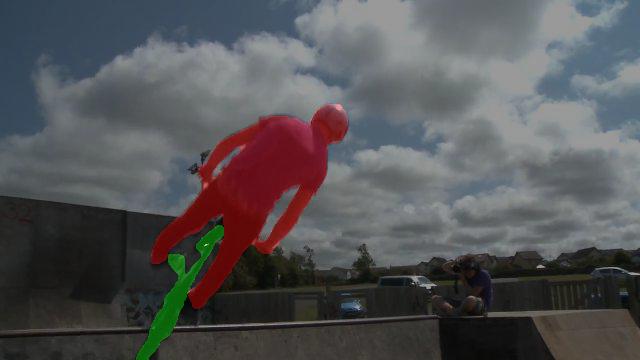}&
		\includegraphics[width=0.195\linewidth]{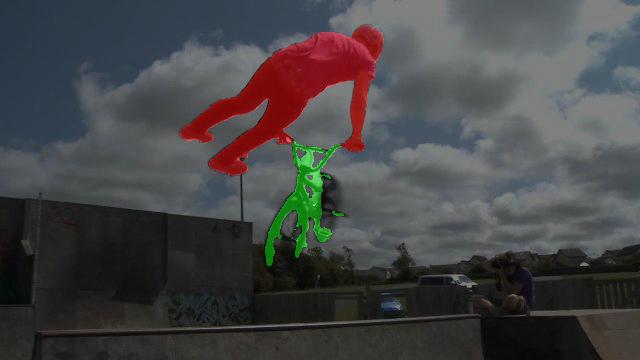}&
		\includegraphics[width=0.195\linewidth]{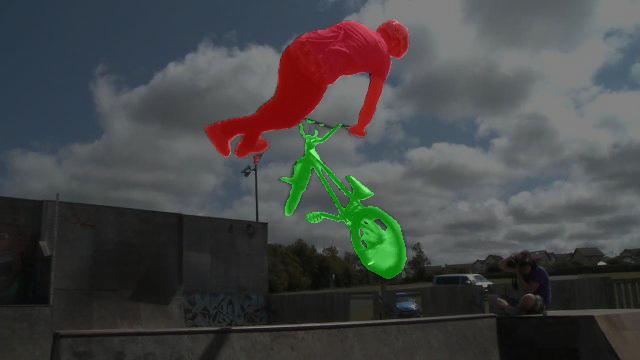}&
		\includegraphics[width=0.195\linewidth]{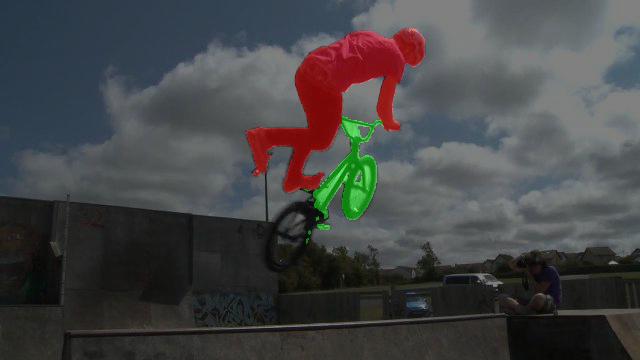}&
		\includegraphics[width=0.195\linewidth]{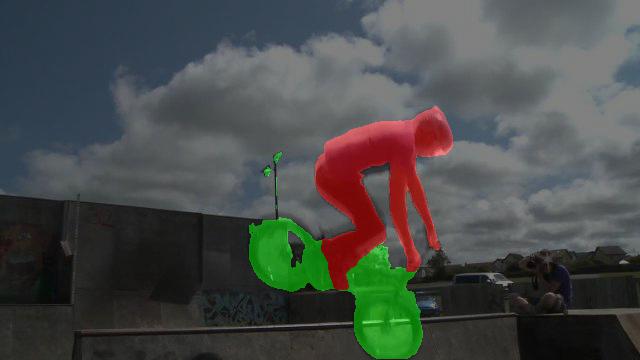}\\
		{\footnotesize (a) 0\% } &{\footnotesize (b) 25\%} &{\footnotesize (c) 50\%}&{\footnotesize (d) 75\%}& {\footnotesize (e) 100\% } \\[0.7mm]
	\end{tabular}
	\caption{Visualization of our results on YouTube-VOS, DAVIS 2017 and SegTrack v2 at different time steps (percentage \wrt the whole video length). The three rows correspond to the YouTube-VOS, DAVIS 2017 and SegTrack v2 datasets respectively.} 
	\vspace{-0.5cm}
	\label{fig:vis_supp}
\end{figure*}

\begin{figure*}[t]
	\centering
	\begin{tabular}{@{\hspace{0mm}}c@{\hspace{0 mm}}c@{\hspace{0 mm}}c@{\hspace{0mm}}c}
		\includegraphics[width=0.25\linewidth]{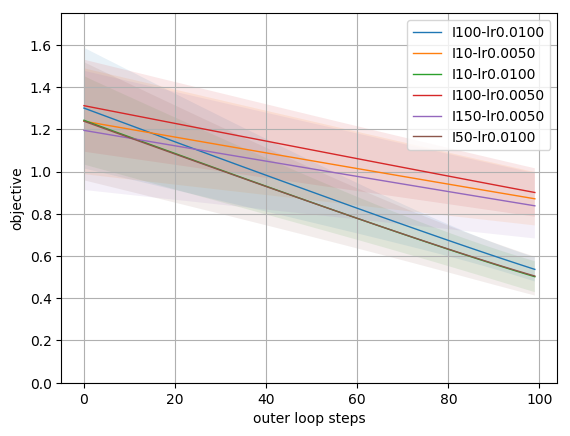}&
		\includegraphics[width=0.25\linewidth]{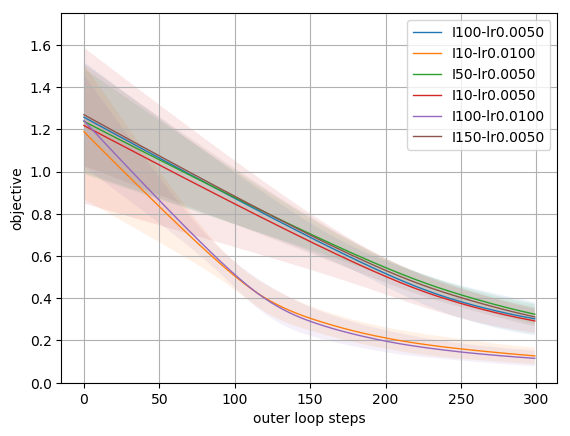}&
		\includegraphics[width=0.25\linewidth]{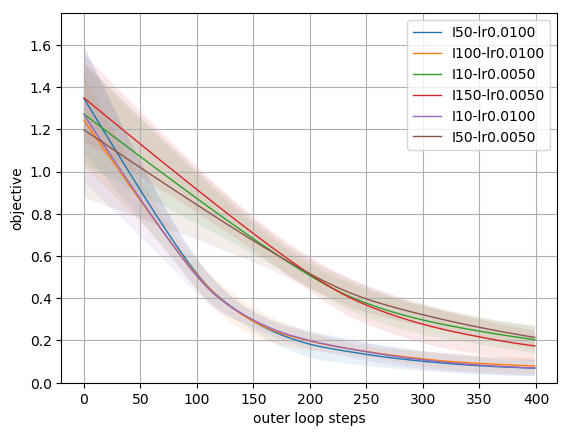}&
		\includegraphics[width=0.25\linewidth]{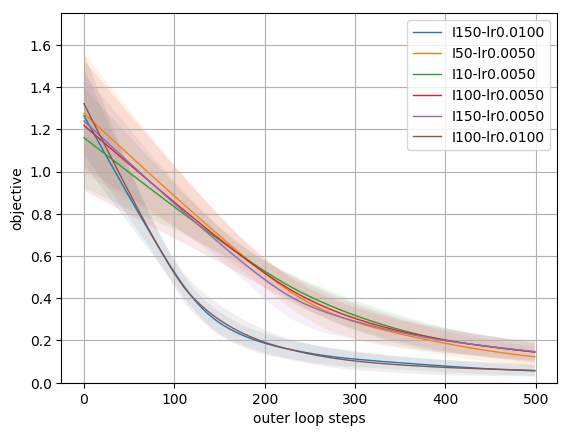}\\[-1.0mm]

		\includegraphics[width=0.25\linewidth]{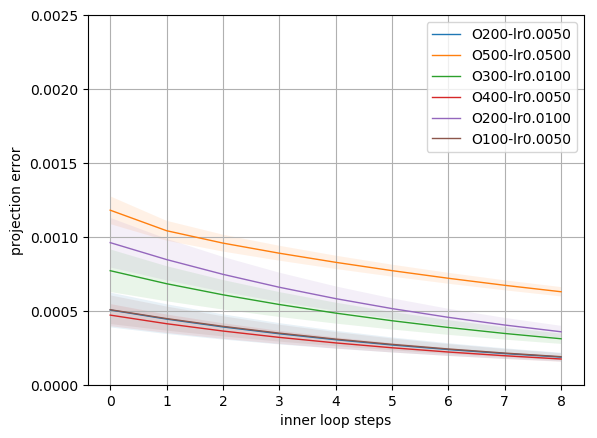}&
		\includegraphics[width=0.25\linewidth]{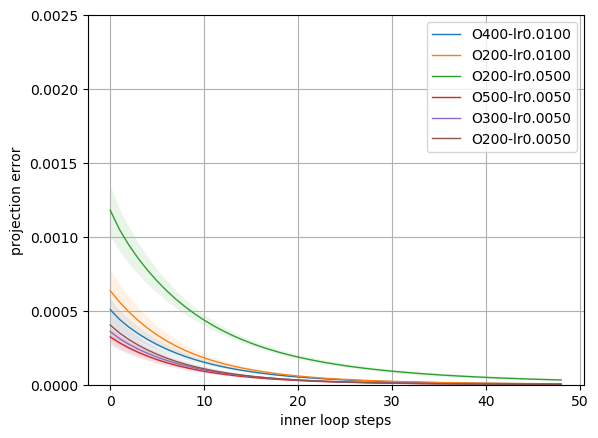}&
		\includegraphics[width=0.25\linewidth]{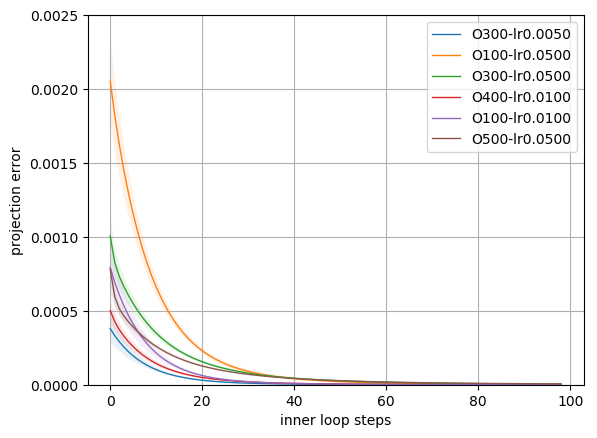}&
		\includegraphics[width=0.25\linewidth]{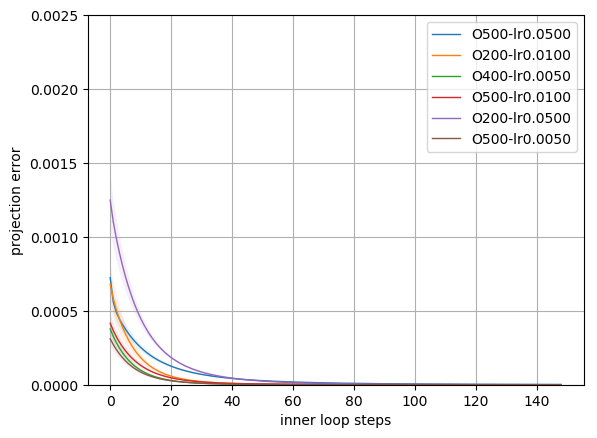}\\[-1.0mm]
		
		\includegraphics[width=0.25\linewidth]{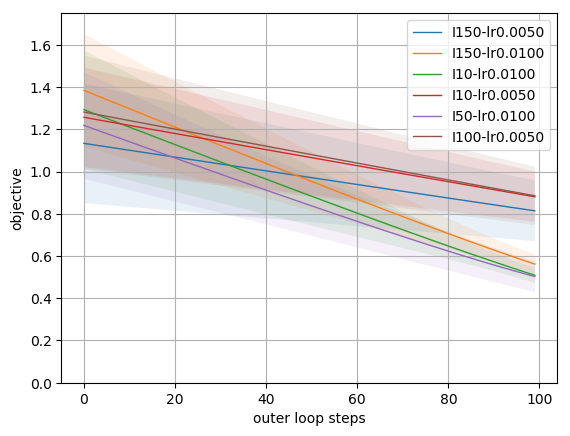}&
		\includegraphics[width=0.25\linewidth]{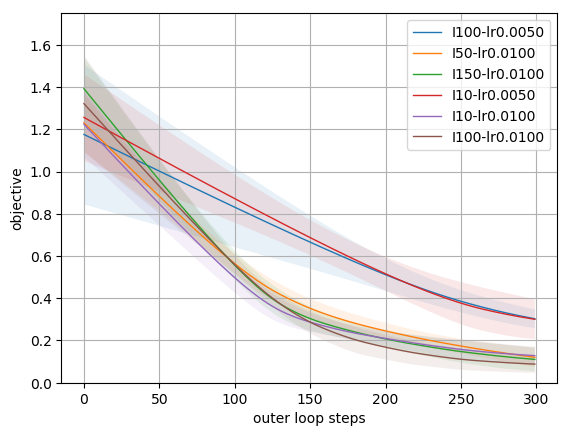}&
		\includegraphics[width=0.25\linewidth]{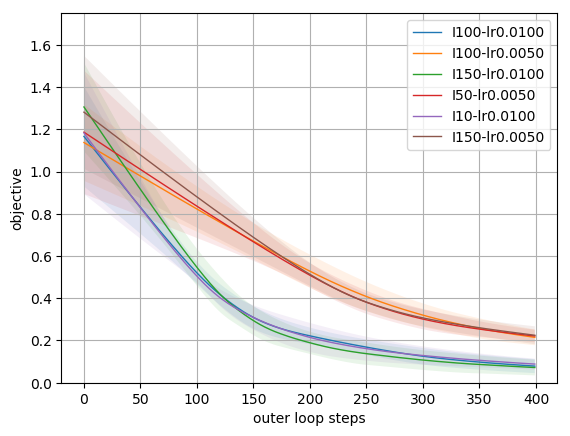}&
		\includegraphics[width=0.25\linewidth]{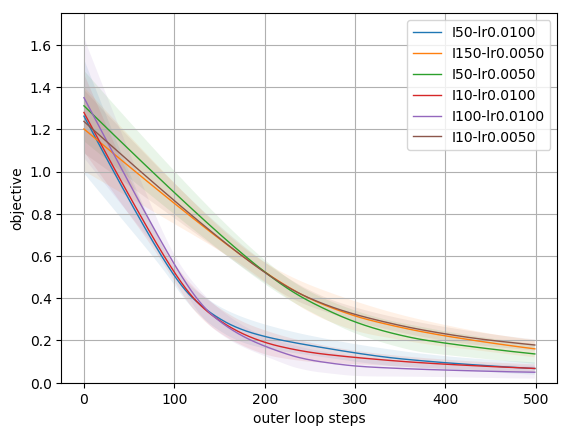}\\[-1.0mm]

		\includegraphics[width=0.25\linewidth]{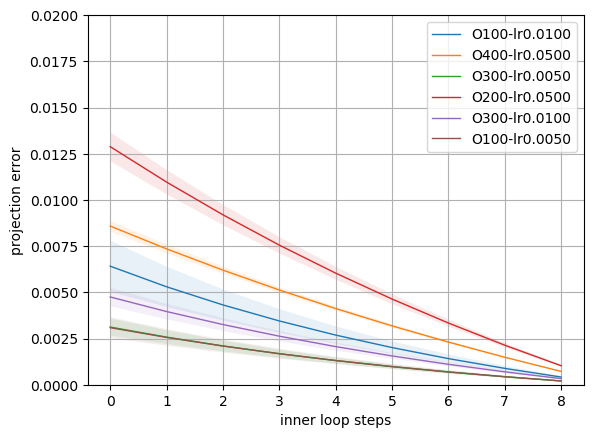}&
		\includegraphics[width=0.25\linewidth]{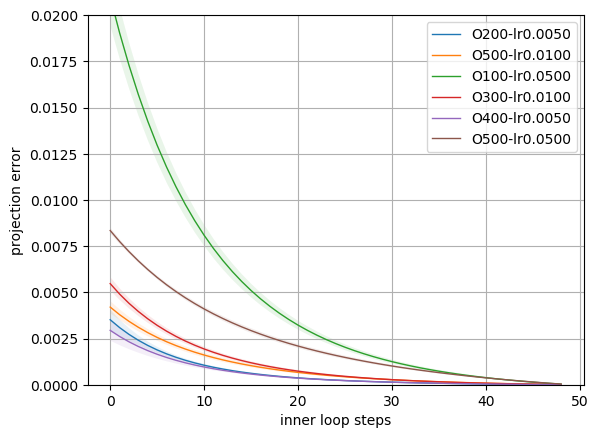}&
		\includegraphics[width=0.25\linewidth]{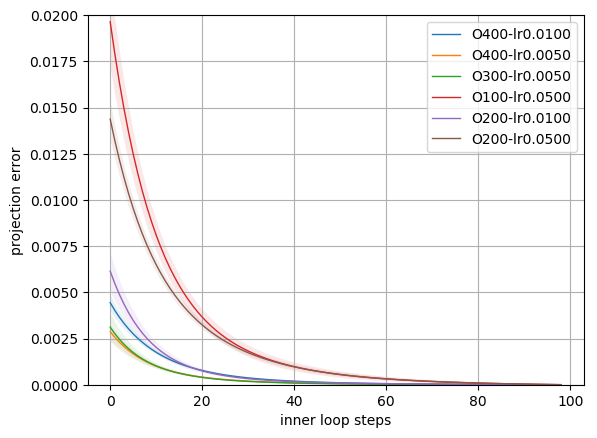}&
		\includegraphics[width=0.25\linewidth]{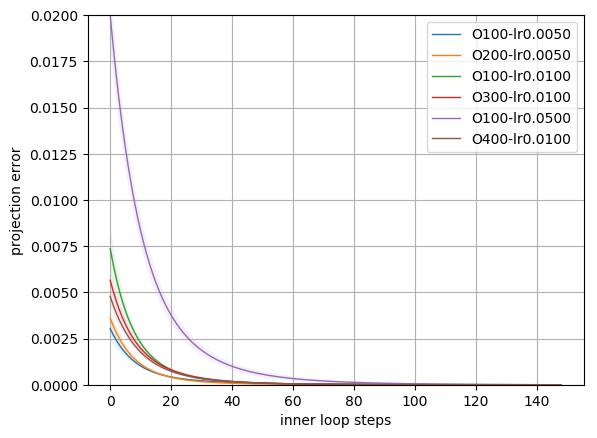}\\[-1.0mm]
		
		\includegraphics[width=0.25\linewidth]{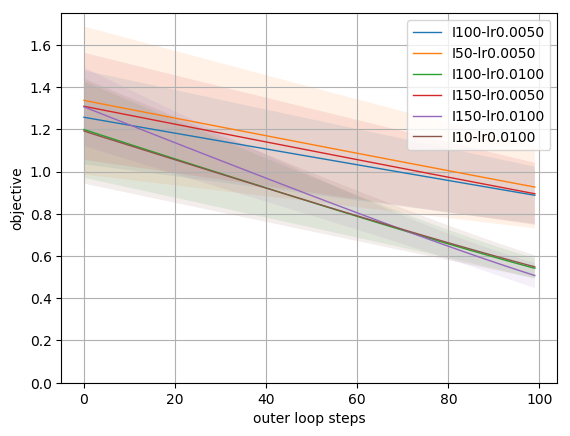}&
		\includegraphics[width=0.25\linewidth]{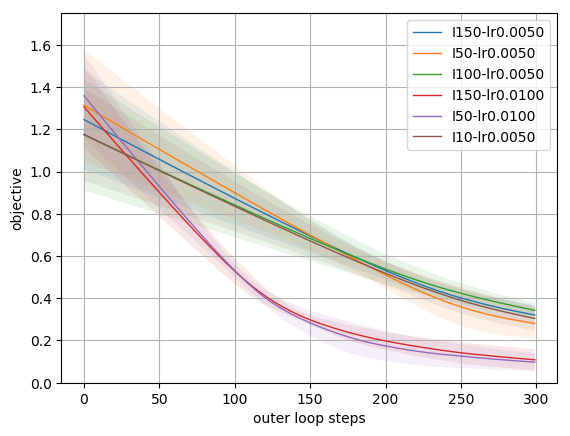}&
		\includegraphics[width=0.25\linewidth]{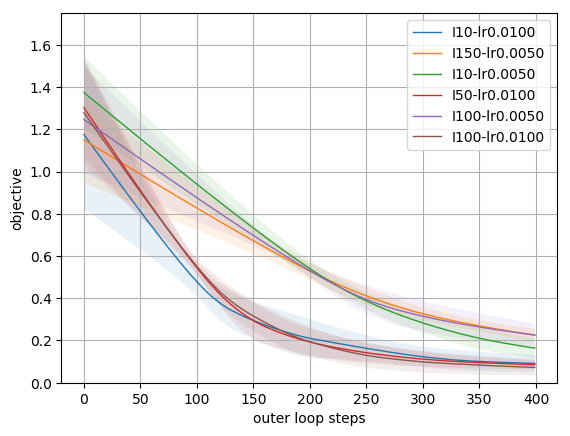}&
		\includegraphics[width=0.25\linewidth]{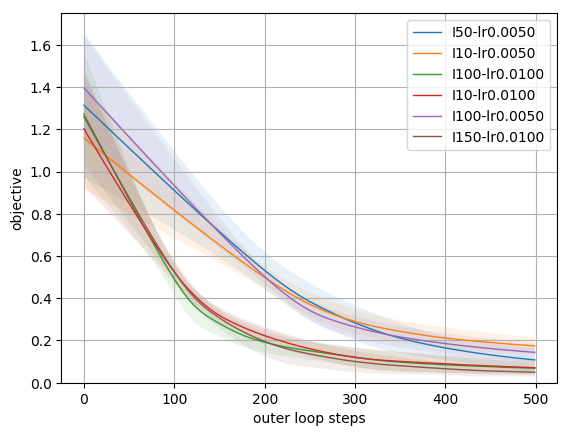}\\[-1.0mm]
		
		\includegraphics[width=0.25\linewidth]{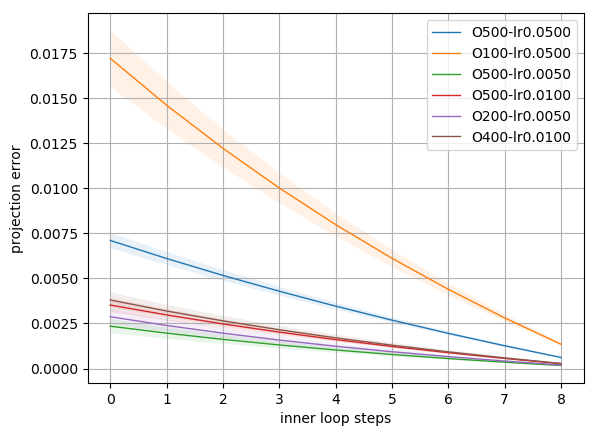}&
		\includegraphics[width=0.25\linewidth]{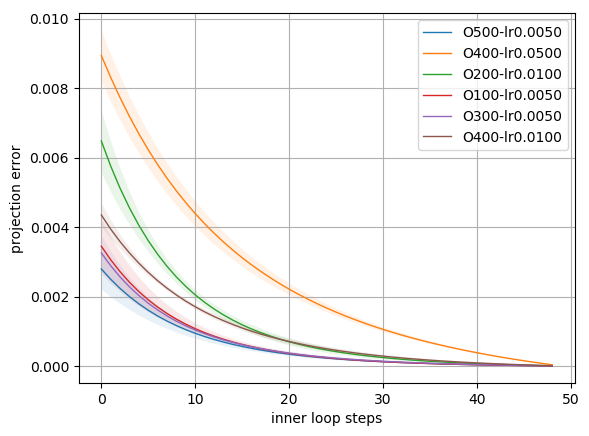}&
		\includegraphics[width=0.25\linewidth]{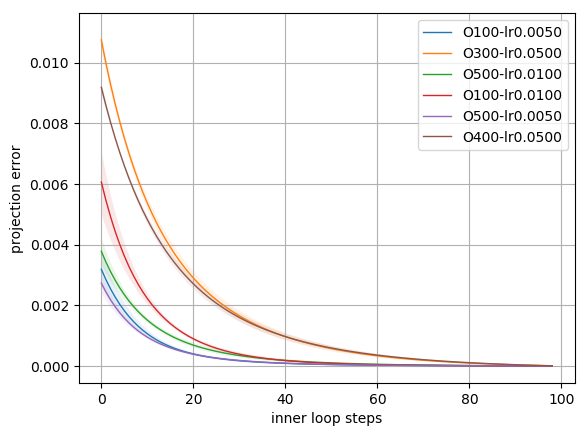}&
		\includegraphics[width=0.25\linewidth]{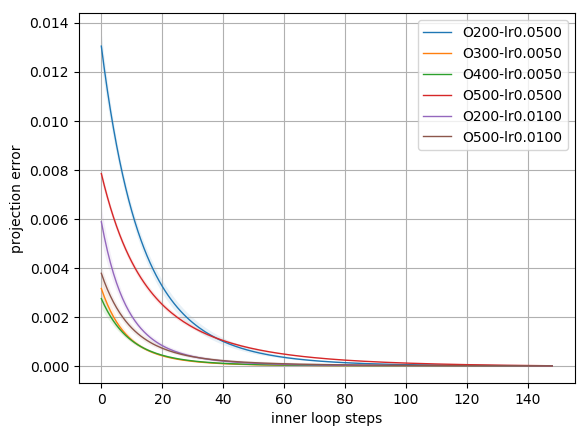}\\[-1.0mm]

	\end{tabular}
	\caption{Hyperparameters (number of steps of outer loop $N_{\text{grad}}$, number of steps of inner loop $N_{\text{proj}}$, and learning rate  $\alpha$) tuning of our differentiable matching layer. 
	We test with three random cost matrices, each of which corresponds to two consecutive rows in the figure, \eg, $1$-st and $2$-nd rows correspond to the $1$-st cost matrix.
    Within the two consecutive rows, the first one shows how the objective function varies with the number of steps in outer loop. 
    The second one shows how projection error varies with the number of steps in inner loops. 
	For each setting of hyperparameters, we perform $10$ different random initialization of the assignment matrix and plot the mean and variance of the curves.}
	\label{fig:hyper-parameters}
\end{figure*}

We show the hyperparameters of our matching algorithm on random cost matrices in Fig.~\ref{fig:hyper-parameters}. The random cost matrices have 5 rows and 100 columns, where the values are sampled from a uniform distribution over $[0, 1)$. Three random cost matrices are generated for the experiment. 
We plot the objective function value of the outer loop and the projection error of the inner loop under different sets of hypperparameters including number of steps of outer loop $N_{\text{grad}}$, number of steps of inner loop $N_{\text{proj}}$ and the learning rate $\alpha$. 
We choose the configuration which gives the best trade-off between the computational cost and the convergent rate: $N_{\text{grad}} = 400$, $N_{\text{proj}} = 50$, and $\alpha=0.01$. From the figure, one can see that the objective value decreases as the number of outer loops increases and it reaches the minimum at about $N_{grad} = 400$ under different setting and different cost matrices. The inner projection error drops in a faster rate and in most of the case it is able to reach almost zero after about 50 rounds of projection. 
The figure also shows that the learning rate plays an important role for the optimization, \ie, learning rate $0.01$ generally leads to a faster convergence than learning rate $0.005$. 
For the experiments on video object segmentation dataset, we start with $N_{\text{grad}} = 400$, $N_{\text{proj}} = 50$ for both training and inference. However, we reduce them to $40$ and $5$, respectively, for the trade off of performance and speed. 

We show additional visual examples of our method on three datasets in Fig.~\ref{fig:vis_supp}.

\section{Example Code}

We show the example PyTorch code (less than $50$ lines) of our differentiable matching algorithm as below.

{
\renewcommand*\familydefault{\ttdefault} 
\begin{lstlisting}[language=Python, float=*b, caption=PyTorch example code]
def project_row(X):
    """
        p(X) = X - 1/m (X 1m - 1n) 1m^T
        X shape: n x m
    """
    X_row_sum = X.sum(dim=1, keepdim=True) # shape n x 1
    one_m = torch.ones(1, X.shape[1]).to(X.device) # shape 1 x m
    return X - (X_row_sum - 1).mm(one_m) / X.shape[1]

def project_col(X):
    """
        p(X) = X  if X^T 1n <= 1m else X - 1/n 1n (1n^T X - 1m^T)  
        X shape: n x m
    """
    X_col_sum = X.sum(dim=0, keepdim=True)     # shape 1 x m
    one_n = torch.ones(X.shape[0], 1).to(X.device) # shape n x 1
    mask = (X_col_sum <= 1).float()
    P = X - (one_n).mm(X_col_sum - 1) / X.shape[0]
    return X * mask + (1 - mask) * P   
        
def relax_matching(C, X_init, max_iter, proj_iter, lr):
    X = X_init 
    P = [torch.zeros_like(C) for _ in range(3)]
    X_list = [X_init]
    
    for i in range(max_iter):               
        X = X - lr * C # gradient step
        
        # project C onto the constrain set 
        for j in range(proj_iter):
            X = X + P[0]
            Y = project_row(X)
            P[0] = X - Y

            X = Y + P[1]
            Y = project_col(X)
            P[1] = X - Y

            X = Y + P[2]
            Y = F.relu(X) 
            P[2] = X - Y

            X = Y
            
        X_list += [X]
    return torch.stack(X_list, dim=0).mean(dim=0) 
\end{lstlisting}
}

\end{document}